\documentclass[10pt,twocolumn,letterpaper]{article}

\usepackage{cvpr}      
\definecolor{cvprblue}{rgb}{0.21,0.49,0.74}
\usepackage[pagebackref,breaklinks,colorlinks,allcolors=cvprblue]{hyperref}

\usepackage{subcaption}
\newtheorem{theorem}{\textbf{Theorem}}

\newtheorem{assumption}{\textbf{Assumption}}

\usepackage{blindtext}
\usepackage{tikz}
\usepackage{threeparttable}
\usepackage[ruled,linesnumbered]{algorithm2e}
\usepackage{diagbox}
\usepackage{multirow}
\usepackage{algpseudocode}
\usepackage{color}
\usepackage{amssymb}
\usepackage{amsmath}
\usepackage[table]{xcolor}
\usepackage{booktabs}

\usepackage{framed}
\definecolor{formalshade}{rgb}{0.95,0.95,0.97}
\definecolor{darkblue}{rgb}{0.14,0.22,0.52}

\def\paperID{16736} 
\def\confName{CVPR}
\def\confYear{2026}

\title{Decoupling Defense Strategies for Robust Image Watermarking}

\author{Jiahui Chen$^1$, Zehang Deng$^2$, Zeyu Zhang$^3$, Chaoyang Li$^1$, Lianchen Jia$^1$, Lifeng Sun$^{1, 4, *}$  \\
$^1$Tsinghua University, $^2$ Swinburne University of Technology,
$^3$ The Australian National University\\
$^4$Key Laboratory of Pervasive Computing, Ministry of Education, $^*$ Corresponding Author\\
\texttt{chenjiah22@mails.tsinghua.edu.cn, sunlf@tsinghua.edu.cn}
}

\begin{document}
\maketitle
\begin{abstract}
Deep learning-based image watermarking, while robust against conventional distortions, remains vulnerable to advanced adversarial and regeneration attacks. Conventional countermeasures, which jointly optimize the encoder and decoder via a noise layer, face 2 inevitable challenges:
(1) decrease of clean accuracy due to decoder adversarial training and (2) limited robustness due to simultaneous training of all three advanced attacks.
To overcome these issues, we propose AdvMark, a novel two-stage fine-tuning framework that decouples the defense strategies. In stage 1, we address adversarial vulnerability via a tailored adversarial training paradigm that primarily fine-tunes the encoder while only conditionally updating the decoder. This approach learns to move the image into a non-attackable region, rather than modifying the decision boundary, thus preserving clean accuracy.
In stage 2, we tackle distortion and regeneration attacks via direct image optimization. To preserve the adversarial robustness gained in stage 1, we formulate a principled, constrained image loss with theoretical guarantees, which balances the deviation from cover and previous encoded images. We also propose a quality-aware early-stop to further guarantee the lower bound of visual quality.
Extensive experiments demonstrate AdvMark outperforms with the highest image quality and comprehensive robustness, i.e. up to 29\%, 33\% and 46\% accuracy improvement for distortion, regeneration and adversarial attacks, respectively.
\end{abstract}    
\section{Introduction}
\label{sec_introduction}
The advances of powerful generative models, such as Stable Diffusion \cite{rombach2022high} and Sora \cite{sora}, necessitate robust mechanisms for tracing and authenticating AI-generated content (AIGC). Deep learning-based watermarking has emerged as a promising solution, embedding a secret message within an image that a corresponding decoder can later extract. 
To mitigate transmission distortions like JPEG compression, previous works \cite{fang2022pimog} leverage joint adversarial optimization (JAT) on both encoder and decoder via a noise layer to simulate various attacks, as depicted in case 2 in Fig.~\ref{figure_2b}.
While effective against common attacks, recent diffusion-based image regeneration \cite{zhao2023invisible} and more advanced adversarial attacks like WEvade \cite{jiang2023evading} have manifested much stronger evasion performance.
We identify that naive joint training paradigm is inherently vulnerable due to 2 critical issues:

\begin{figure}
\centering
\includegraphics[width=1.\linewidth]{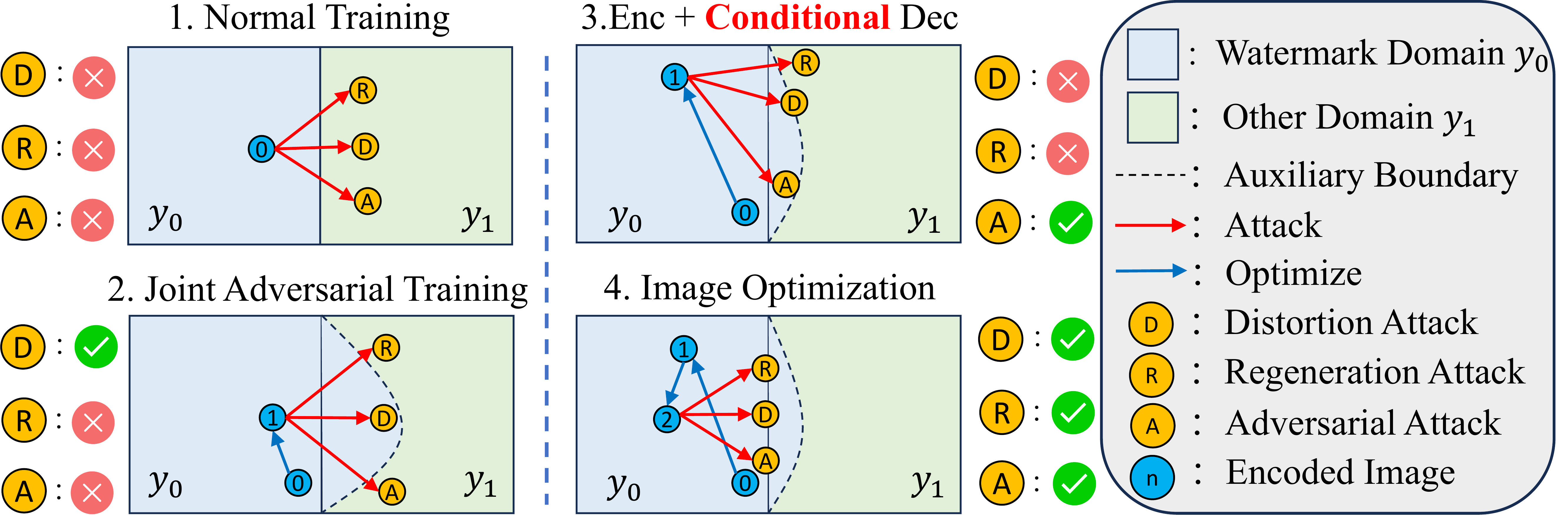}
\caption{Four training paradigms. Adversarial training (2) degrades clean accuracy of $y_1$ and exhibits limited improvement, while moving image (3+4) suffices in both terms.}
\label{figure_2b}
\end{figure}

\begin{figure}
\centering
\includegraphics[width=0.83\linewidth]{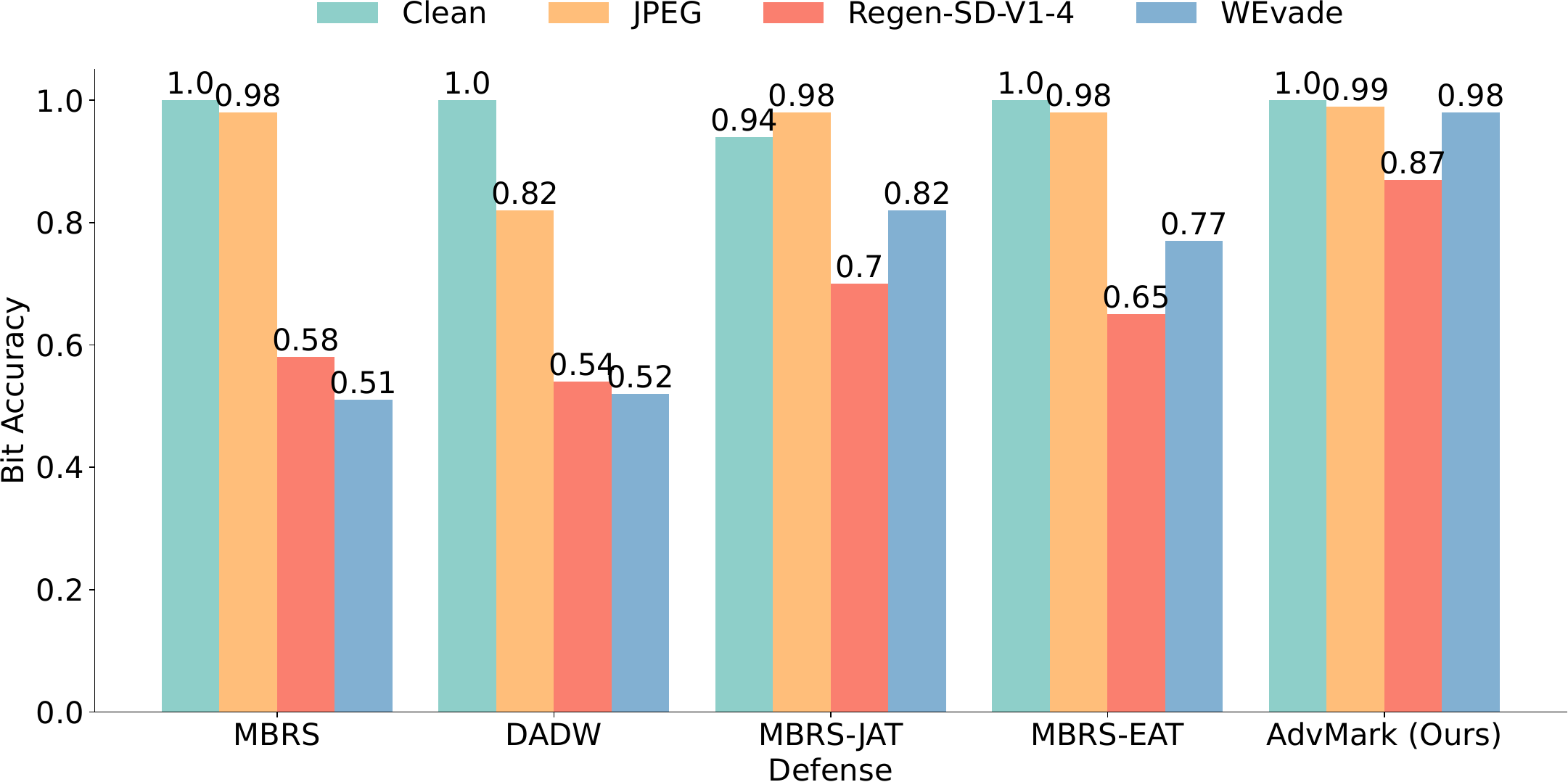}
\caption{The bit accuracy $\uparrow$ of against three representative distortion, regeneration and adversarial attacks. JAT and EAT denote joint and encoder-based adversarial training.}
\label{figure_2a}
\end{figure}

\textbf{Challenge 1}: \textit{Decrease of clean accuracy.}
It is well understood that decoder adversarial training gains robustness with inevitably lower clean accuracy \cite{zhang2019towards,pang2020bag} on unattacked images due to tampered boundary distribution.
As a result, JAT optimization also sacrifices accuracy for marginal robustness on each attack.
We present an empirical experiment in Fig.~\ref{figure_2a} (refer to Section \ref{subsec_setup} for setup).
Without robustness training, MBRS and DADW both exhibit low accuracy against Regen and WEvade.
Applying JAT to MBRS successfully improves robustness but the clean accuracy instead decreases to 0.94 due to significant landscape modification, as shown in case 2 in Fig.~\ref{figure_2b}.
To guarantee both accuracy, our insight is to harness the power of encoder fine-tuning (EAT) to move the images towards the center (case 3 in Fig.~\ref{figure_2b}), instead of expanding auxiliary boundary.
As shown in Fig.~\ref{figure_2a}, MBRS-EAT maintains high clean accuracy while achieving competitive robustness compared to MBRS-JAT.

\textbf{Challenge 2}: \textit{Simultaneous training yields limited robustness.}
While encoder-focused adversarial training (EAT) solves the clean accuracy problem, it is not a panacea. 
The accuracy against Regen (0.65) and WEvade (0.77) still exhibits gaps from JPEG. This is because these attacks have more intricate mechanisms, e.g. numerous sampling iterations of the diffusion model and derivatives from the decoder. Forcing a single model to simultaneously defend against all attacks within a monolithic training process leads to an inefficient and slow-converging optimization.
To offload the training, our insight is to separate the defense stage with an additional direct image optimization on distortion and regeneration attacks (case 4 in Fig.~\ref{figure_2b}), while we address the adversarial attack only via EAT due to two reasons:
(1) it requires first-order derivatives of the image, and optimization on attacked image incurs higher-order derivatives which incur prohibitive computational and memory overhead;
(2) past work has observed that ReLU neural networks are locally almost linear \cite{ioffe2015batch}, which suggests that second derivatives may be close to zero in most cases and makes optimization hard to converge \cite{finn2017model}.
By decoupling the defense, we can achieve the highest clean and robust accuracy, as shown in Fig.~\ref{figure_2a}.

Built upon the insights, we propose AdvMark, a framework that abandons the joint training paradigm in favor of a decoupling, two-stage fine-tuning process. 
In stage 1, we propose a novel adversarial training paradigm that focuses on encoder fine-tuning. This is guided by an improved defender-side adversarial attack construction. We optimize the message to deviate from the ground-truth rather than towards a random label.
The decoder is only conditionally updated when robustness is below the threshold, ensuring that clean accuracy is not compromised.
While in stage 2, we address other attacks via efficient and effective direct image optimization.
To maintain the robustness obtained in stage 1, we propose a novel constrained image loss to not only enhance visual quality, but also limit the deviation from previously encoded image.
The performance is guaranteed by a theorem based on practical and verifiable assumption.
To further guarantee quality, we propose to improve upon the known PGD \cite{madry2017towards} optimization with a quality-aware early-stop, rather than the implicit $\epsilon$-ball projection.

We conduct extensive experiments with 9 watermarking methods against 10 attacks to demonstrate that, AdvMark outperforms with up to 46\% accuracy improvement and the highest image quality in terms of PSNR, SSIM and LPIPS.

We summarize our contributions as follows:

$\bullet$
To our best knowledge, we are the first to evaluate existing watermarking against distortion, regeneration, and adversarial attacks systematically.
We demonstrate that typical joint optimization exhibits 2 challenges:
(1) low clean accuracy due to decoder training and (2) limited robustness due to simultaneous training of non-trivial attacks.
Then we derive two key insights through empirical evaluation.

$\bullet$
To bridge the gap, we propose AdvMark, a two-stage comprehensively robust image watermarking method.
In stage 1, we propose an improved defender tailored adversarial attack, accompanied by a novel adversarial training paradigm that mainly fine-tunes the encoder.
The decoder is only conditionally trained to guarantee clean accuracy.
In stage 2, we leverage image optimization to address the distortion and regeneration attacks.
To preserve adversarial robustness, we design a constrained image loss with theoretical guarantees.
We also propose a quality-aware PGD to improve visual quality.

$\bullet$
Extensive experiments indicate the comprehensive robustness of AdvMark with the highest image quality.
The ablation study further validates the efficiency of AdvMark.
\section{Related Work}
\label{sec_background}
\subsection{Image Watermarking}
\label{subsec_deep_watermarking}
Image watermarking has long been studied in the context of intellectual property protection.
Existing watermarking methods can be categorized into three types:
(1) Traditional watermarking methods that leverage heuristic hand-crafted embedding like DwtDctSvd \cite{cox2007digital} in frequency domain.
(2) Post-processing deep learning based alternatives \cite{ma2022towards,fang2022pimog,luo2020distortion} such as HiDDeN \cite{zhu2018hidden} and MBRS \cite{jia2021mbrs}. 
They leverage a noise layer between encoder and decoder to simulate various attack such as JPEG to enhance robustness.
More recent VINE \cite{lurobust} leverages pretrained text-to-image diffusion model to defend against image editing attacks \cite{zhang2024editguard,zhang2025omniguard}.
(3) In-processing methods \cite{wen2024tree,arabihidden} that directly embed watermarks as part of the image generation process.
For example, Stable Signature \cite{fernandez2023stable} fine-tunes the decoder of the LDM to incorporate the message in latent features, leaving the diffusion component unchanged.

\newcommand*\emptycirc[1][1ex]{\tikz\draw (0,0) circle (#1);} 
\newcommand*\halfcirc[1][1ex]{%
\begin{tikzpicture}
\draw[fill] (0,0)-- (90:#1) arc (90:270:#1) -- cycle ;
\draw (0,0) circle (#1);
\end{tikzpicture}}
\newcommand*\fullcirc[1][1ex]{\tikz\fill (0,0) circle (#1);} 
\begin{figure*}
\centering
\includegraphics[width=0.87\linewidth]{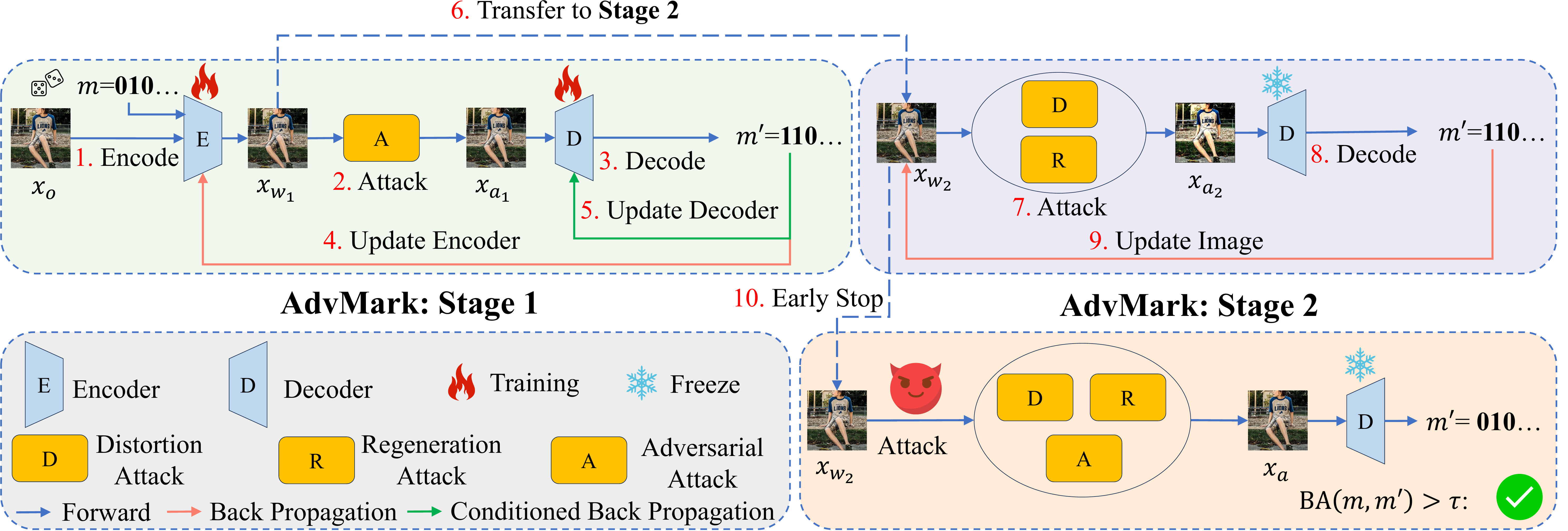}
\caption{Overview of AdvMark. In stage 1 we mainly fine-tune the encoder with slight decoder training to tackle adversarial attack. In stage 2 we directly optimize the encoded image to address the rest two attacks while preserving adversarial robustness.}
\label{figure_3}
\end{figure*}

\subsection{Threat Model And Attacks}
\label{subsec_malicious_attacks}
\noindent\textbf{The defender} produces a watermarked image $x_w = E(x_o, m)$ from clean image $x_o$ and message $m$ via proprietary encoder $E$. The decoder $D$ is subjected to API or model exposure.
\noindent\textbf{The attacker} aims to construct perturbed image $x_a$ such that the decoder fails to recover the correct message (i.e., $D(x_a) \neq m$), while ensuring $x_a$ remains visually similar to $x_w$. The encoder cannot be accessed but the decoder's knowledge varies: (1) white-box with full access (e.g. the model leakage from third party \cite{meta}) or (2) black-box with only query capability.
We consider three types of attacks:

\noindent\textbf{Distortion Attack.}
We evaluate against heuristic transformations like JPEG compression, Gaussian noise, and various geometric attacks, which resemble real-world transmission loss.

\noindent\textbf{Regeneration Attack.}
Given the remarkable generation performance from diffusion models \cite{zhao2023invisible,saberi2023robustness}, we can destroy the watermark with semantic content reconstruction. Specifically, it first destructs a watermarked image by adding noise to its representation (the forward process) and then removes the watermark pattern via efficient denoising process like DDIM \cite{song2020denoising}.

\noindent\textbf{Adversarial Attack.}
For white-box with full access, we can construct malicious images to evade decoder $D$ with minimal, often imperceptible modifications like WEvade \cite{jiang2023evading}. It crafts an adversarial example such that the decoded message from $D$ approaches a random target message $m_t$ within a perturbation budget $r$, formulated as:
\begin{equation}
\label{equ_wevade}
    \min_{\delta} \ \ l(D(x_w+\delta),m_t), \ s.t. \Vert \delta\Vert_{\infty }<r
\end{equation}

where $l$ denotes the $L_2$ loss and $x_w$ is the watermarked image.
In the \textbf{black-box} setting, a query-based attack \cite{jiang2023evading} perturbs and adjusts the image based on feedback from the decoder's output.
Alternatively, \citet{saberi2023robustness} proposes a transfer-based attack by training a surrogate decoder on watermarked images. Adversarial examples are then crafted and transferred to evade the target decoder. 

\section{Proposed Method: AdvMark}
\label{sec_method}
\subsection{Overview}
\label{subsec_overview}

Built upon previous insights, we propose the two-stage framework overview of AdvMark in Fig.~\ref{figure_3}.
In stage 1, we propose a novel encoder-based fine-tuning strategy to tackle adversarial attack.
We encode the original image $x_o$ and derive the attacked $x_{a_1}$ via our defender tailored adversarial attack.
The decoded message $m_{a_1}$ incurs $L_1$ loss.
After several rounds of fine-tuning on encoder, we conditionally update the decoder if the bit accuracy of $m_{a_1}$ is below a prefixed threshold.
While in stage 2, we directly optimize the encoded image $x_{w_2}$ to preserve visual quality and previous adversarial robustness.
Finally, we leverage a quality-aware metric to early stop the optimization and obtain the encoded image, which is capable of defending against all three kinds of attacks without degrading quality or clean accuracy.

\subsection{Stage 1: Adversarial Encoder Fine-tuning}
\label{subsec_stage1}
\textbf{Adversarial Examples.}
We first fine-tune the pretrained encoder to handle adversarial attack.
We denote original image $x_o \in \mathbb{R}^{3\times H \times W} $, $H,W$ are height and width of an image, a secret message of length $n$ is $m \in \{0,1\}^n$, the encoded image from encoder $E$ is $x_{w_1}=E(x_o,m)\in \mathbb{R}^{3\times H \times W}$, while the extracted message logits after decoder $D$ is $y_{w_1}=D(x_{w_1}) \in \mathbb{R}^n$, and $m_{w_1}=clamp(round(y_{w_1}),0,1)$.
The final bit accuracy $BA(m,m_{w_1})$ is the fraction of matched bits.
To construct adversarial examples $x_{a_1}=x_{w_1}+\delta$, we improve optimization in Equ. \ref{equ_wevade} and propose a defender tailored loss.
Since we have the ground-truth secret message $m$, we can directly render $m_{a_1}$ random to $m$ (i.e. $BA(m_{a_1},m)\rightarrow 0.5$):
\begin{small}
\begin{equation}
\label{equ_white_attack}
    \min_{\delta} \ \ |0.5-l(clamp(D(x_{w_1}+\delta),0,1),m)|, \ s.t. \Vert \delta\Vert_{\infty }<r
\end{equation}
\end{small}

\begin{algorithm}
    \caption{Stage 1 of AdvMark}
    \label{algorithm_1}
    \KwIn{pretrained Encoder $\theta_E$ and Decoder $\theta_D$, image datasets $I$, other parameters $iter_E$, $\alpha_E$, $\alpha_D$, $r$, $\lambda_{w_1}$, $\lambda_{i_1}$ and $\tau_1$}
    \KwOut{fine-tuned Encoder $\theta_E^*$ and Decoder $\theta_D^*$} 
    \For {batched images $x_o$ in $I$} 
    {
    sample random message $m \in \{0,1\}^n$;\\
    \For{$i \leftarrow 1$ to $iter_E$}
    {
    $x_{w_1}\leftarrow E(x_o,m)$; $x_{a_1}\leftarrow x_{w_1}+\delta^*(r)$;\\
    $L_1\leftarrow L_{a_1}+\lambda_{w_1}L_{w_1}+\lambda_{i_1}L_{i_1}$ in Equ.~6;\\
    $\theta_E\leftarrow \theta_E-\alpha_E \cdot \frac{\partial L_1}{\partial \theta_E}$;\\
    \If{$i==iter_E$}{
    $m_{a_1}\leftarrow clamp(round(D(x_{a_1})),0,1)$;\\
    \lIf{$BA(m_{a_1},m)<\tau_1$}{$\theta_D\leftarrow \theta_D-\alpha_D \cdot \frac{\partial L_1}{\partial \theta_D}$}
    }
    }
    }
    \textbf{return} $\theta_E,\theta_D$
\end{algorithm}

where $l$ denotes the mean-squared-error (MSE) loss throughout the paper, and $r$ is the perturbation budget to maintain image quality.
$clamp(D(x_{w_1}+\delta),0,1)$ represent the bit possibility of $m_{a_1}$ aligned with $m$.
Note that we do not optimize $m_{a_1}$ due to zero gradient from $round(*)$ operation.
The final $\delta^*$ is derived via PGD \cite{madry2017towards} optimization.

\noindent\textbf{Optimization Loss.}
With adversarial example $x_{a_1}$, we tend to minimize the adversarial loss such that $m_{a_1}$ approaches $m$ under budget $r$, formulated as $L_{a_1}$:
\begin{equation}
\label{equ_La_1}
    L_{a_1}=l(D(x_{a_1}),m),\ x_{a_1}=x_{w_1}+\delta^*
\end{equation}
To maintain clean accuracy we derive $L_{w_1}$ formulated as:
\begin{equation}
\label{equ_Lw_1}
    L_{w_1}=l(D(x_{w_1}),m)
\end{equation}
To enhance image visual quality, we leverage the MSE distance and LPIPS \cite{zhang2018unreasonable} loss to guarantee both pixel and semantic similarity, formulated as:
\begin{equation}
\label{equ_Li_1}
    L_{i_1}=(l(x_{w_1},x_o)+LPIPS(x_{w_1},x_o))/2
\end{equation}
Together we have the total loss as:
\begin{equation}
\label{equ_L1}
L_1=L_{a_1}+\lambda_{w_1}L_{w_1}+\lambda_{i_1}L_{i_1}
\end{equation}
where $\lambda_{w_1}$ and $\lambda_{i_1}$ denote the weights for clean accuracy and image quality, respectively.

\begin{figure}
\centering
\includegraphics[width=1.0\linewidth]{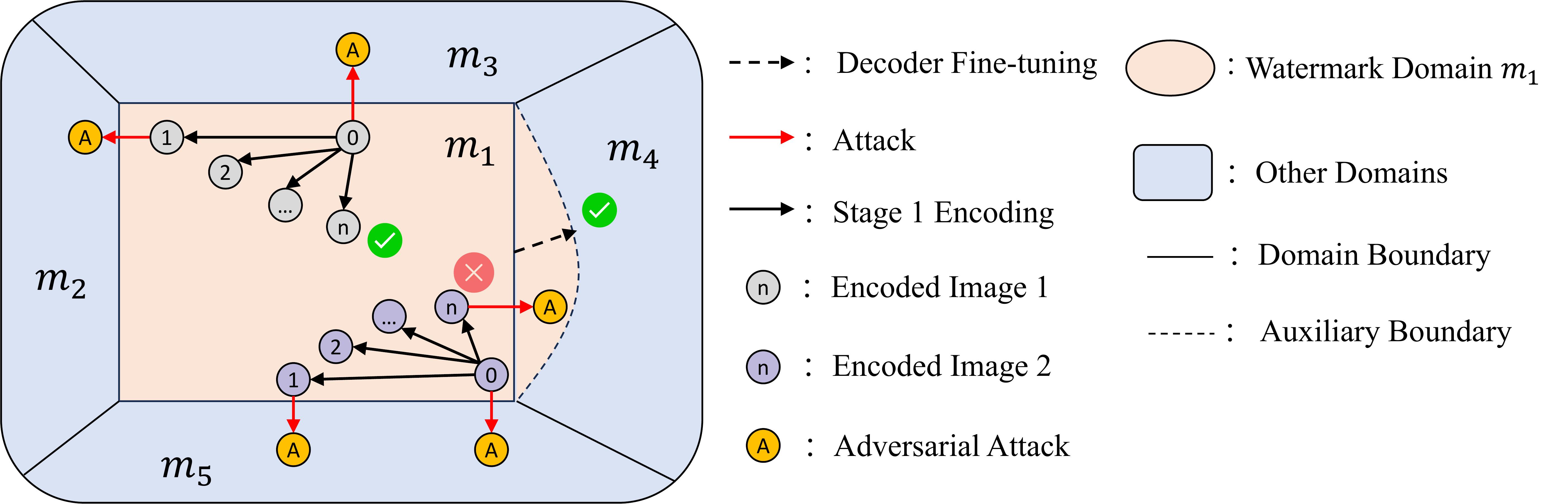}
\caption{Illustration of stage 1. We constantly fine-tune the encoder to map the image (No. 1) towards the non-attackable center (No. n). The decoder is updated only when the final image fails to suffice (No. n of image 2).}
\label{figure_10a}
\end{figure}
\noindent\textbf{Fine-tuning.}
We propose to mainly fine-tune the encoder for $iter_E$ rounds.
If the final robustness accuracy $BA(m_{a_1},m)$ is below a fixed threshold $\tau_1$, the decoder is updated once with $L_1$ loss.
We present the algorithm of Stage 1 of AdvMark in Algorithm \ref{algorithm_1}.

\noindent\textbf{Example.}
We illustrate stage 1 in Fig.~\ref{figure_10a}.
The encoder is fine-tuned to mitigate adversarial attack by moving the image into a safe zone.
If the encoded image $n$ remains vulnerable anyway, we slightly update the decoder once to expand auxiliary boundary.

\subsection{Stage 2: Quality-aware Image Optimization}
\label{subsec_stage2}
\textbf{Optimization Loss.}
Following \textbf{Insight 2}, we propose to directly optimize the encoded image $x_{w_1}$ to enhance robustness against distortion and regeneration attacks.
We first derive the attack loss as follows:
\begin{equation}
\label{equ_La_2}
    L_{a_2}=\frac{\sum_{k=2}^{K+2}  \lambda_{a_k} \cdot l(D(x_{a_k}),m)}{\sum_{k=2}^{K+2}  \lambda_{a_k}}
\end{equation}

where $x_{a_k}=x_{w_2}+\delta x_{a_k},\ k \in [2,K+2]$, $K$ denotes the distortion number, $x_{a_{K+2}}$ denotes the regeneration attack and
$\lambda_{a_k}$ denotes optimization weights for different attacks.
Note that for each attack we apply the differentiable implementation to ensure optimization of $x_{a_k}$.
Similar to Equ.~\ref{equ_Lw_1}, we also maintain the clean accuracy of $x_{w_2}$ as follows:
\begin{equation}
\label{equ_Lw_2}
    L_{w_2}=l(D(x_{w_2}),m)
\end{equation}

To preserve the adversarial robustness in stage 1, we propose a novel constrained image loss that additionally includes the deviation distance between $(x_{w_1},x_{w_2})$, which is formulated as:
\begin{equation}
\label{equ_Li_2}
L_{i_2}=\frac{l(x_{w_2},x_o)+LPIPS(x_{w_2},x_o)+2\cdot l(x_{w_2},x_{w_1})}{4} 
\end{equation}

Together the total loss for stage 2 is in Equ.~\ref{equ_L2}, where $\lambda_{w_2}$ and $\lambda_{i_2}$ represent weights for clean accuracy and image quality, respectively.
\begin{equation}
\label{equ_L2}
L_2=L_{a_2}+\lambda_{w_2}L_{w_2}+\lambda_{i_2}L_{i_2}
\end{equation}

To demonstrate the effectiveness of $l(x_{w_2},x_{w_1})$ on robustness preservation, we propose theorem \ref{theorem} based on an assumption.
\begin{algorithm}
    \caption{Stage 2 of AdvMark}
    \label{algorithm_2}
    \KwIn{original image $x_o$, fine-tuned encoder $E$, message $m$, other parameters $iter_o$, $\alpha_x$, $p$, $\lambda_{w_2}$, $\lambda_{i_2}$, $\lambda_{a_k},k\in [2,K+2]$, $\tau_2$}
    \KwOut{full-scale robust images $x_{w_2}$}
    $x_{w_2}\leftarrow E(x_o,m)$;\\
    \For{$i \leftarrow 1$ to $iter_o$}
    {
    \For{$k$ in $[2\sim K+2]$}
    {
    $x_{a_k}\leftarrow x_{w_2}+\delta x_{a_k}$;\\
    $m_{a_k}\leftarrow clamp(round(D(x_{a_k})),0,1)$;\\
    \lIf{$BA(m_{a_k},m)\ge \tau_2$}{del $x_{a_k}$}
    }
    $L_2\leftarrow L_{a_2}+\lambda_{w_2}L_{w_2}+\lambda_{i_2}L_{i_2}$ in Equ.~10;\\
    $x_{w_2}\leftarrow I(x_{w_2}-\alpha_x \cdot sign(\frac{\partial L_2}{\partial x_{w_2}}) )$;\\
    \lIf{$PSNR(x_{w_2},x_o)\le p$}{break}
    }
    \textbf{return} $x_{w_2}$
\end{algorithm}

\begin{figure}
\centering
\includegraphics[width=1.0\linewidth]{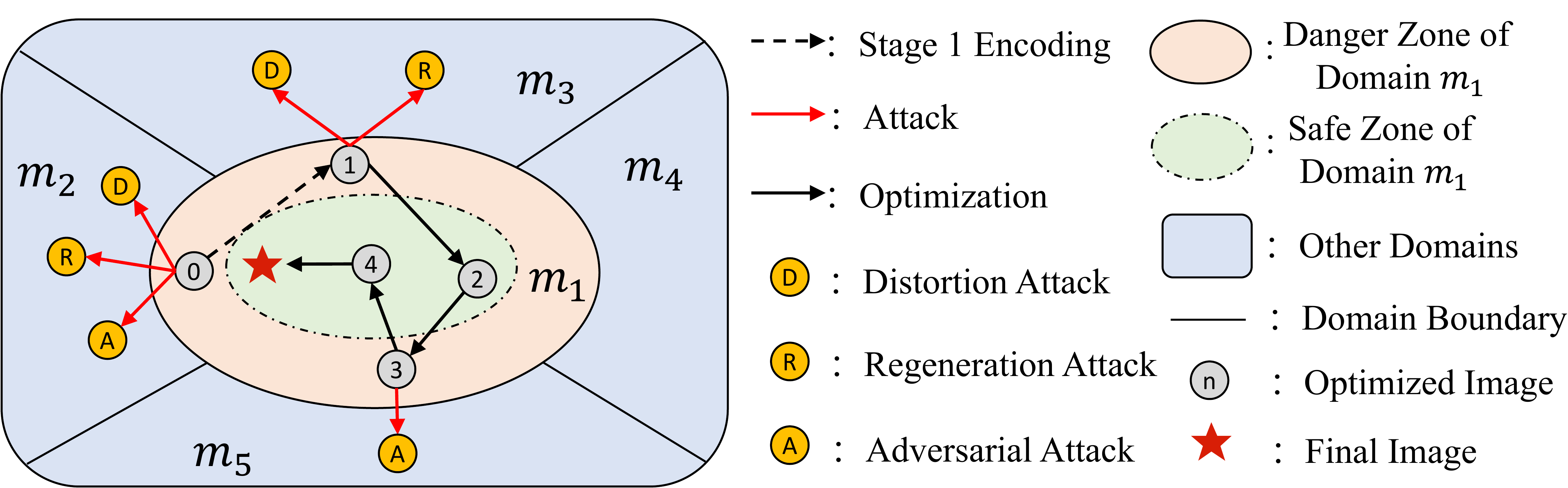}
\caption{Illustration of stage 2. 
$0 \rightarrow 1$: initialize image 1 from encoder, which exhibits only adversarial robustness;
2: comprehensive robustness but low visual quality;
3: high quality yet vulnerable to A attack again;
4: similar to image 2;
$\star$: high quality and robustness.}
\label{figure_10b}
\end{figure}

\begin{assumption}
\label{ass_1}
Assume the constrained image loss Equ. \ref{equ_Li_2} is minimized such that $\Vert x_{w_2}-x_{w_1} \Vert\le \delta\ <\alpha$. 
\end{assumption}

\begin{theorem}
\label{theorem}
Given a robust $x_{w_1}$, i.e. $f(x_{w_1})=f(x_{w_1}+\eta_1)$,$\forall \ \Vert\eta_1 \Vert\le\alpha$, where $f$ maps images into messages.
Let assumption \label{ass_1} hold, then $x_{w_2}$ is also robust with an adjusted budget, i.e. $f(x_{w_2})=f(x_{w_2}+\eta_2)$,$\forall \ \Vert\eta_2 \Vert\le \alpha-\delta$.
\end{theorem}

\noindent\textbf{Empirical Results}.
We verify the assumption with $L_2$: $\alpha=0.012$, $\delta=0.007$ and the actual $\eta_2$ bound is $0.010>\alpha-\delta$. The technical proof is trivial because based on assumption \ref{ass_1}, we have $f(x_{w_1})=f(x_{w_2})$, while we also have $\Vert x_{w_2}+\eta_2-x_{w_1}\Vert \le \Vert x_{w_2}-x_{w_1}\Vert+\Vert \eta_2\Vert\le \delta+\alpha-\delta=\alpha$, hence $f(x_{w_2}+\eta_2)=f(x_{w_1})=f(x_{w_2})$.

\noindent\textbf{Direct Optimization.}
To minimize $L_2$ loss, we improve upon the PGD optimization with quality-aware mapping as follows:
\begin{equation}
\label{equ_pgd}
    x_{w_2}=I(x'_{w_2},x_{w_2}),x'_{w_2}=x_{w_2}-\alpha_x \cdot sign(\frac{\partial L_2}{\partial x_{w_2}}) 
\end{equation}
\begin{equation}
\label{equ_map}
    I(x'_{w_2},x_{w_2}) =
\begin{cases}
x'_{w_2} & \text{if } PSNR(x'_{w_2},x_o) \ge p, \\
x_{w_2} & \text{if } PSNR(x'_{w_2},x_o) < p.
\end{cases}
\end{equation}

where $\alpha_x$ is the learning rate.
We replace the typical $\epsilon$-ball projection $\prod_{x_o,\epsilon}$ with $I(*,*)$ mapping. 
In this way, we directly constrain the lower bound of visual quality via budget $p$.
During the optimization, we disregard attacks in $L_{a_2}$ whose bit accuracy of $x_{a_k}$ is above the threshold $\tau_2$ to guarantee comprehensive robustness.
Finally we early stop the optimization if PSNR is below budget $p$.
The summarized steps for stage 2 are in Algorithm \ref{algorithm_2}.



\noindent\textbf{Example.}
We illustrate stage 2 in Fig.~\ref{figure_10b}.
The initial image 1 gains only adversarial robustness from vulnerable image 0.
During $1\rightarrow2$ we optimize $L_{a_2}$ to enhance overall robustness, which comes with low visual quality and clean accuracy.
Hence we further optimize $L_{i_2}+L_{w_2}$ to obtain image 3, which again remains vulnerable to adversarial attack.
During $3\rightarrow4$ we optimize $l(x_{w_2},x_{w_1})$ to approach image 1 and gain previous robustness.
Finally, we optimize $L_{i_2}$ to further enhance image quality while maintaining comprehensive robustness.
\section{Experiments}
\label{sec_evaluation}
\subsection{Experimental Setup}
\label{subsec_setup}

\begin{figure*}
\centering
\includegraphics[width=0.93\linewidth]{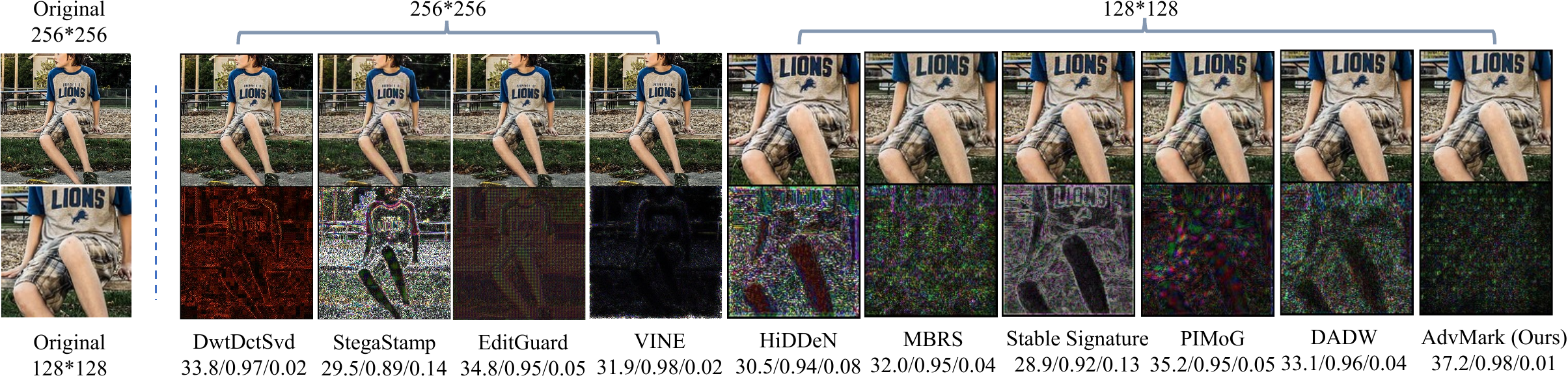}
\caption{The original $x_o$, watermarked $x_w$ and residual images $x_r$ of different watermarking methods where $x_r=|x_w-x_o|\times 10$.
We present below each watermarking method as PSNR$\uparrow$/SSIM$\uparrow$/LPIPS$\downarrow$.}
\label{figure_4}
\end{figure*}

\begin{table}
\centering
\caption{Visual quality of watermarked images, n is the secret message length. \colorbox[HTML]{ECF4FF}{Blue} and \colorbox[HTML]{FFE5E3}{Red} denote the best and worst.}
\resizebox{\linewidth}{!}
{\begin{tabular}{c|c|c|ccc|ccc}
    \toprule[1.6pt]
    \multirow{2}{*}{Defense} & \multirow{2}{*}{Image Size} & \multirow{2}{*}{n} & \multicolumn{3}{c|}{MS-COCO} & \multicolumn{3}{c}{DiffusionDB} \\ \cmidrule{4-9} 
     &  &  & PSNR$\uparrow$ & SSIM$\uparrow$ & LPIPS$\downarrow$ & PSNR$\uparrow$ & SSIM$\uparrow$ & LPIPS$\downarrow$ \\ \midrule
    DwtDctSvd & 256 $\times$ 256 & 30 & 35.8 & 0.98 & 0.03 & 36.8 & 0.98 & 0.05 \\
    HiDDeN & 128 $\times$ 128 & 30 & 30.3 & 0.94 & 0.14 & \colorbox[HTML]{FFE5E3}{30.8} & \colorbox[HTML]{FFE5E3}{0.94} & \colorbox[HTML]{FFE5E3}{0.16} \\
    MBRS & 128 $\times$ 128 & 30 & 32.1 & 0.95 & 0.09 & 31.8 & \colorbox[HTML]{FFE5E3}{0.94} & 0.14 \\
    Stable Signature & 128 $\times$ 128 & 48 & 30.5 & 0.94 & 0.12 & \colorbox[HTML]{FFE5E3}{30.8} & \colorbox[HTML]{FFE5E3}{0.94} & 0.14 \\
    PIMoG & 128 $\times$ 128 & 30 & 35.8 & 0.95 & 0.08 & 35.6 & 0.95 & 0.10 \\
    DADW & 128 $\times$ 128 & 30 & 33.4 & 0.98 & 0.05 & 33.6 & 0.98 & 0.04 \\
    StegaStamp & 256 $\times$ 256 & 64 & \colorbox[HTML]{FFE5E3}{29.8} & \colorbox[HTML]{FFE5E3}{0.93} & \colorbox[HTML]{FFE5E3}{0.15} & 31.7 & \colorbox[HTML]{FFE5E3}{0.94} & 0.14 \\
    EditGuard & 256 $\times$ 256 & 64 & 36.3 & 0.95 & 0.04 & 36.1 & \colorbox[HTML]{FFE5E3}{0.94} & 0.04 \\
    VINE & 256 $\times$ 256 & 100 & 35.5 & 0.98 & \colorbox[HTML]{ECF4FF}{0.01} & 35.2 & \colorbox[HTML]{ECF4FF}{0.99} & \colorbox[HTML]{ECF4FF}{0.01} \\ \midrule
    \multirow{4}{*}{AdvMark (Ours)} & 128 $\times$ 128 & 30 & 37.0 & \colorbox[HTML]{ECF4FF}{0.99} & \colorbox[HTML]{ECF4FF}{0.01} & 36.9 & \colorbox[HTML]{ECF4FF}{0.99} & \colorbox[HTML]{ECF4FF}{0.01} \\
     & 128 $\times$ 128 & 48 & 37.2 & \colorbox[HTML]{ECF4FF}{0.99} & \colorbox[HTML]{ECF4FF}{0.01} & 37.0 & \colorbox[HTML]{ECF4FF}{0.99} & \colorbox[HTML]{ECF4FF}{0.01} \\
     & 256 $\times$ 256 & 64 & 38.4 & \colorbox[HTML]{ECF4FF}{0.99} & \colorbox[HTML]{ECF4FF}{0.01} & 38.3 & \colorbox[HTML]{ECF4FF}{0.99} & \colorbox[HTML]{ECF4FF}{0.01} \\
     & 256 $\times$ 256 & 100 & \colorbox[HTML]{ECF4FF}{38.9} & \colorbox[HTML]{ECF4FF}{0.99} & \colorbox[HTML]{ECF4FF}{0.01} & \colorbox[HTML]{ECF4FF}{38.8} & \colorbox[HTML]{ECF4FF}{0.99} & \colorbox[HTML]{ECF4FF}{0.01} \\ \midrule[1.6pt]
    \end{tabular}}
\label{table_psnr}
\end{table}

All our experiments are conducted on NVIDIA RTX 4090.
\noindent\textbf{Datasets and Metrics.}
We evaluate AdvMark on MS-COCO \cite{lin2014microsoft} and DiffusionDB \cite{wang2022diffusiondb}.
For image visual quality, we adopt the well-known PSNR, Structural Similarity score (SSIM) \cite{wang2004image}, and semantic metric Learned Perceptual Image Patch Similarity (LPIPS) \cite{zhang2018unreasonable}.
For watermarking performance, we use the bit accuracy $BA(m_1,m_2)$, i.e. the fraction of matched bits.

\noindent\textbf{Parameter Setup.}
For distortion attacks, we follow previous work in \cite{jia2021mbrs,jiang2023evading} and adopt their default setting.
For regeneration attacks, we adopt 2 Stable Diffusion checkpoints for image reconstruction.
Adversarial attacks include WEvade and the black-box query-based (Black-Q) from \cite{jiang2023evading} with target threshold $\tau'=0.75$.
We also follow \cite{saberi2023robustness} for surrogate-based attack (Black-S).

For all watermarking methods, we set both thresholds $\tau_1=\tau_2=0.95$ throughout the paper.
For AdvMark, we fine-tune the pretrained MBRS models \cite{jia2021mbrs} with the following parameters:
In stage 1, encoder fine-tuning iteration $iter_E=10$, learning rate $\alpha_E=\alpha_D=5e^{-4}$, adversarial attack budget $r=\frac{20}{255}$, $\lambda_{w_1}=1$, $\lambda_{i_1}=3$.
While in stage 2, we consider $K=4$ distortion attacks and set JPEG weight $\lambda_{a_2}=1$, the rest $\lambda_{a_k}=0.1,k\in [3,6]$, optimization iteration $iter_o=10$, learning rate $\alpha_x=5e^{-2}$, loss weight $\lambda_{w_2}=0.1$ and $\lambda_{i_2}=5$, PSNR budget $p=36$.
Due to inconsistent image size and message length among baselines, we also train MBRS with different setup.

Regarding baselines, we include 7 methods:
traditional scheme DwtDctSvd \cite{cox2007digital}, learning-based HiDDeN \cite{zhu2018hidden}, MBRS \cite{jia2021mbrs}, PIMoG \cite{fang2022pimog}, StegaStamp \cite{tancik2020stegastamp}, DADW \cite{luo2020distortion}, EditGuard \cite{zhang2024editguard} (degrade version), VINE \cite{lurobust} (robust version) and in-processing method Stable Signature \cite{fernandez2023stable}.
To ensure a fair comparison, we fine-tune all the baselines via typical joint optimization with all attacks, except for EditGuard and VINE due to their specific design in tampering detection and editing defense.

\begin{table}
\centering
\caption{Robust accuracy$\uparrow$. $1,2,3,4,1\sim4$ represent JPEG, Gaussian noise, Gaussian blur, brightness and 4 attacks combined respectively. $^*$ denotes unknown attacks not included in training. \textbf{Geometric and advanced combined attacks are in Table \ref{table_ablation_distortion}}.}
\resizebox{1.0\linewidth}{!}
{\begin{tabular}{c|c|c|ccccc|cc|ccc}
\toprule[1.6pt]
 &  &  & \multicolumn{5}{c|}{Distortion} & \multicolumn{2}{c|}{Regeneration} & \multicolumn{3}{c}{Adversarial} \\ \cmidrule{4-13} 
\multirow{-2}{*}{Datasets} & \multirow{-2}{*}{\diagbox[]{Defense}{Attack}} & \multirow{-2}{*}{Clean} & 1 & 2 & 3 & 4 & (1$\sim$4)$^*$ & V1-4 & V1-5$^*$ & WEvade & B-S$^*$ & B-Q$^*$ \\ \midrule
 & DwtDctSvd & 0.99 & 0.88 & 0.96 & 0.99 & \cellcolor[HTML]{FFE5E3}0.60 & \cellcolor[HTML]{FFE5E3}0.54 & 0.62 & 0.63 & / & 0.63 & 0.73 \\
 & HiDDeN & \cellcolor[HTML]{FFE5E3}0.88 & \cellcolor[HTML]{FFE5E3}0.70 & \cellcolor[HTML]{FFE5E3}0.75 & \cellcolor[HTML]{FFE5E3}0.87 & 0.85 & 0.64 & 0.57 & 0.56 & 0.57 & 0.68 & 0.73 \\
 & MBRS & 0.93 & 0.98 & \cellcolor[HTML]{ECF4FF}1.00 & \cellcolor[HTML]{ECF4FF}1.00 & \cellcolor[HTML]{ECF4FF}1.00 & 0.76 & 0.70 & 0.70 & 0.82 & \cellcolor[HTML]{ECF4FF}1.00 & 0.73 \\
 & Stable Signature & 0.90 & 0.80 & 0.88 & 0.91 & 0.94 & 0.70 & 0.65 & 0.65 & 0.78 & 0.87 & 0.73 \\
 & PIMoG & 0.93 & 0.85 & 0.91 & \cellcolor[HTML]{ECF4FF}1.00 & 0.98 & 0.68 & 0.66 & 0.66 & 0.77 & 0.80 & 0.73 \\
 & DADW & \cellcolor[HTML]{ECF4FF}1.00 & 0.82 & 0.88 & \cellcolor[HTML]{ECF4FF}1.00 & 0.98 & 0.58 & \cellcolor[HTML]{FFE5E3}0.54 & \cellcolor[HTML]{FFE5E3}0.54 & 0.52 & \cellcolor[HTML]{FFE5E3}0.55 & 0.73 \\
 & StegaStamp & 0.92 & 0.94 & 0.95 & \cellcolor[HTML]{ECF4FF}1.00 & 0.88 & 0.79 & 0.71 & \cellcolor[HTML]{ECF4FF}0.71 & 0.79 & 0.56 & 0.72 \\
 & EditGuard & 1.00 & 0.82 & \cellcolor[HTML]{FFE5E3}0.75 & 0.94 & 0.95 & 0.61 & 0.59 & 0.59 & \cellcolor[HTML]{FFE5E3}0.48 & 0.75 & 0.73 \\
 & VINE & 1.00 & 0.96 & 0.94 & 0.93 & 0.96 & 0.78 & 0.74 & 0.75 & 0.51 & 0.80 & 0.73 \\
\multirow{-10}{*}{COCO} & AdvMark (Ours) & \cellcolor[HTML]{ECF4FF}1.00 & \cellcolor[HTML]{ECF4FF}0.99 & \cellcolor[HTML]{ECF4FF}1.00 & \cellcolor[HTML]{ECF4FF}1.00 & \cellcolor[HTML]{ECF4FF}1.00 & \cellcolor[HTML]{ECF4FF}0.83 & \cellcolor[HTML]{ECF4FF}0.87 & \cellcolor[HTML]{ECF4FF}0.87 & \cellcolor[HTML]{ECF4FF}0.98 & \cellcolor[HTML]{ECF4FF}1.00 & 0.73 \\ \midrule
 & DwtDctSvd & \cellcolor[HTML]{ECF4FF}1.00 & 0.90 & 0.95 & 0.99 & \cellcolor[HTML]{FFE5E3}0.58 & \cellcolor[HTML]{FFE5E3}0.53 & 0.58 & 0.59 & / & 0.62 & 0.72 \\
 & HiDDeN & \cellcolor[HTML]{FFE5E3}0.86 & \cellcolor[HTML]{FFE5E3}0.70 & \cellcolor[HTML]{FFE5E3}0.72 & \cellcolor[HTML]{FFE5E3}0.85 & 0.84 & 0.64 & 0.56 & 0.57 & 0.55 & 0.65 & 0.73 \\
 & MBRS & 0.91 & 0.98 & \cellcolor[HTML]{ECF4FF}1.00 & \cellcolor[HTML]{ECF4FF}1.00 & \cellcolor[HTML]{ECF4FF}1.00 & 0.72 & 0.69 & 0.69 & 0.80 & \cellcolor[HTML]{ECF4FF}1.00 & 0.73 \\
 & Stable Signature & 0.88 & 0.72 & 0.83 & 0.86 & 0.90 & 0.66 & 0.64 & 0.64 & 0.76 & 0.82 & 0.71 \\
 & PIMoG & 0.92 & 0.84 & 0.90 & \cellcolor[HTML]{ECF4FF}1.00 & 0.97 & 0.66 & 0.62 & 0.62 & 0.73 & 0.82 & 0.72 \\
 & DADW & \cellcolor[HTML]{ECF4FF}1.00 & 0.81 & 0.88 & 1.00 & 0.96 & 0.57 & \cellcolor[HTML]{FFE5E3}0.53 & \cellcolor[HTML]{FFE5E3}0.53 & \cellcolor[HTML]{FFE5E3}0.51 & \cellcolor[HTML]{FFE5E3}0.57 & 0.72 \\
 & StegaStamp & 0.91 & 0.95 & 0.96 & 0.91 & 0.76 & 0.79 & 0.72 & 0.72 & 0.78 & 0.59 & 0.72 \\
 & EditGuard & 1.00 & 0.80 & 0.74 & 0.94 & 0.93 & 0.60 & 0.55 & 0.56 & \cellcolor[HTML]{FFE5E3}0.51 & 0.70 & 0.72 \\
 & VINE & 1.00 & 0.95 & 0.91 & 0.91 & 0.94 & 0.75 & 0.76 & 0.76 & 0.53 & 0.78 & 0.73 \\
\multirow{-10}{*}{DB} & AdvMark (Ours) & \cellcolor[HTML]{ECF4FF}1.00 & \cellcolor[HTML]{ECF4FF}0.98 & \cellcolor[HTML]{ECF4FF}1.00 & \cellcolor[HTML]{ECF4FF}1.00 & \cellcolor[HTML]{ECF4FF}1.00 & \cellcolor[HTML]{ECF4FF}0.83 & \cellcolor[HTML]{ECF4FF}0.85 & \cellcolor[HTML]{ECF4FF}0.85 & \cellcolor[HTML]{ECF4FF}0.96 & \cellcolor[HTML]{ECF4FF}1.00 & 0.74 \\ \midrule[1.6pt]
\end{tabular}}
\label{table_acc}
\end{table}

\subsection{Image Visual Quality}
\label{subsec_quality}
As shown in Table \ref{table_psnr}, the baselines exhibit limited quality due to complex joint training for robustness against all the attacks.
Our method surpasses the original MBRS with significant improvement (32.1 to 37.0), thanks to our two-stage design and optimization in Equ.~\ref{equ_Li_1} and \ref{equ_Li_2}.
To better perceive the visual quality, we present an example to compare the residual images in Fig.~\ref{figure_4}.
Most methods embed watermark inside the background to mitigate noticeable artifacts, while AdvMark outperforms by reducing overall noises to improve visual similarity.

\begin{figure}[]
    \centering
    \begin{subfigure}{0.48\linewidth}
        \centering
        \includegraphics[width=\linewidth]{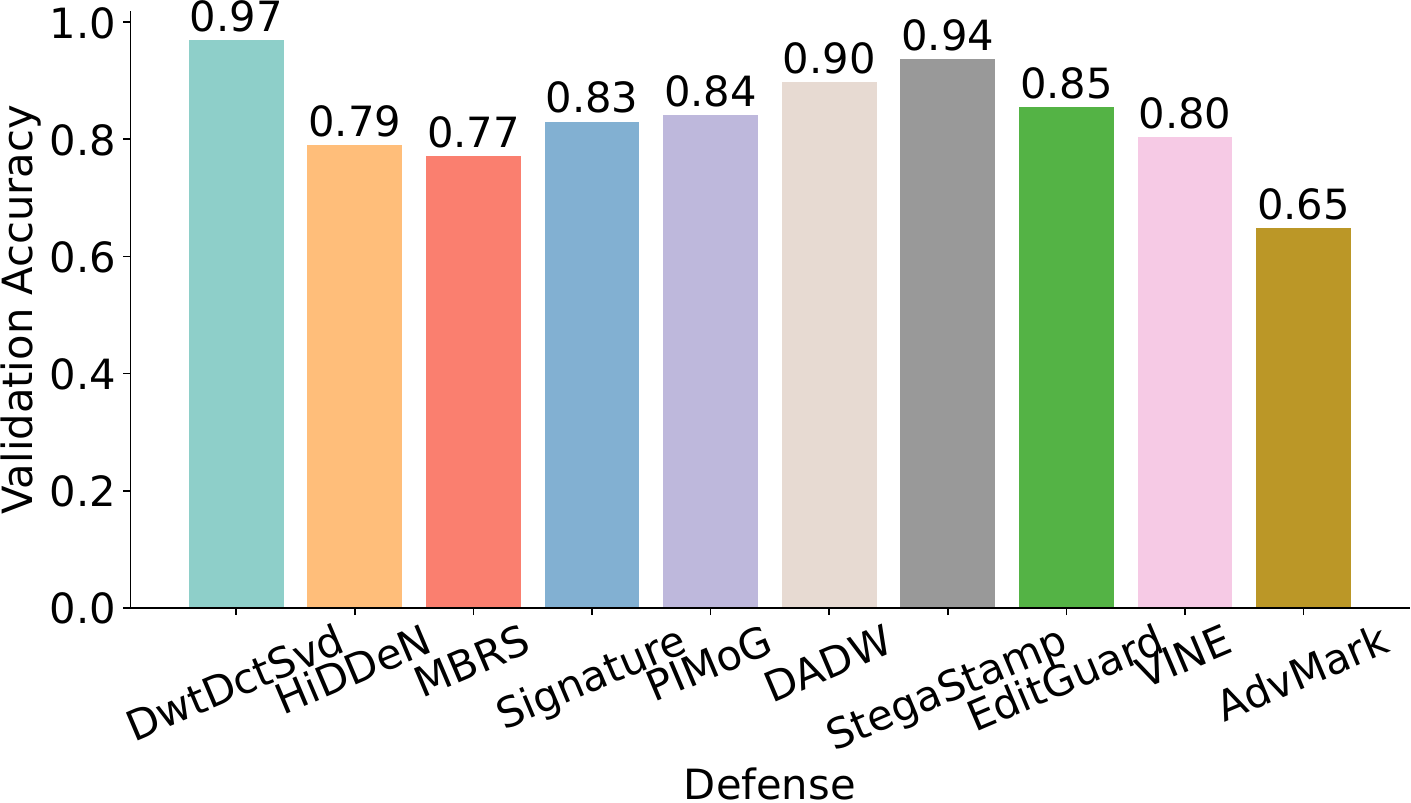}
        \caption{Validation accuracy$\downarrow$.}
        \label{fig:surrogate_accuracy}
    \end{subfigure}
    \begin{subfigure}{0.48\linewidth}
        \centering
        \includegraphics[width=\linewidth]{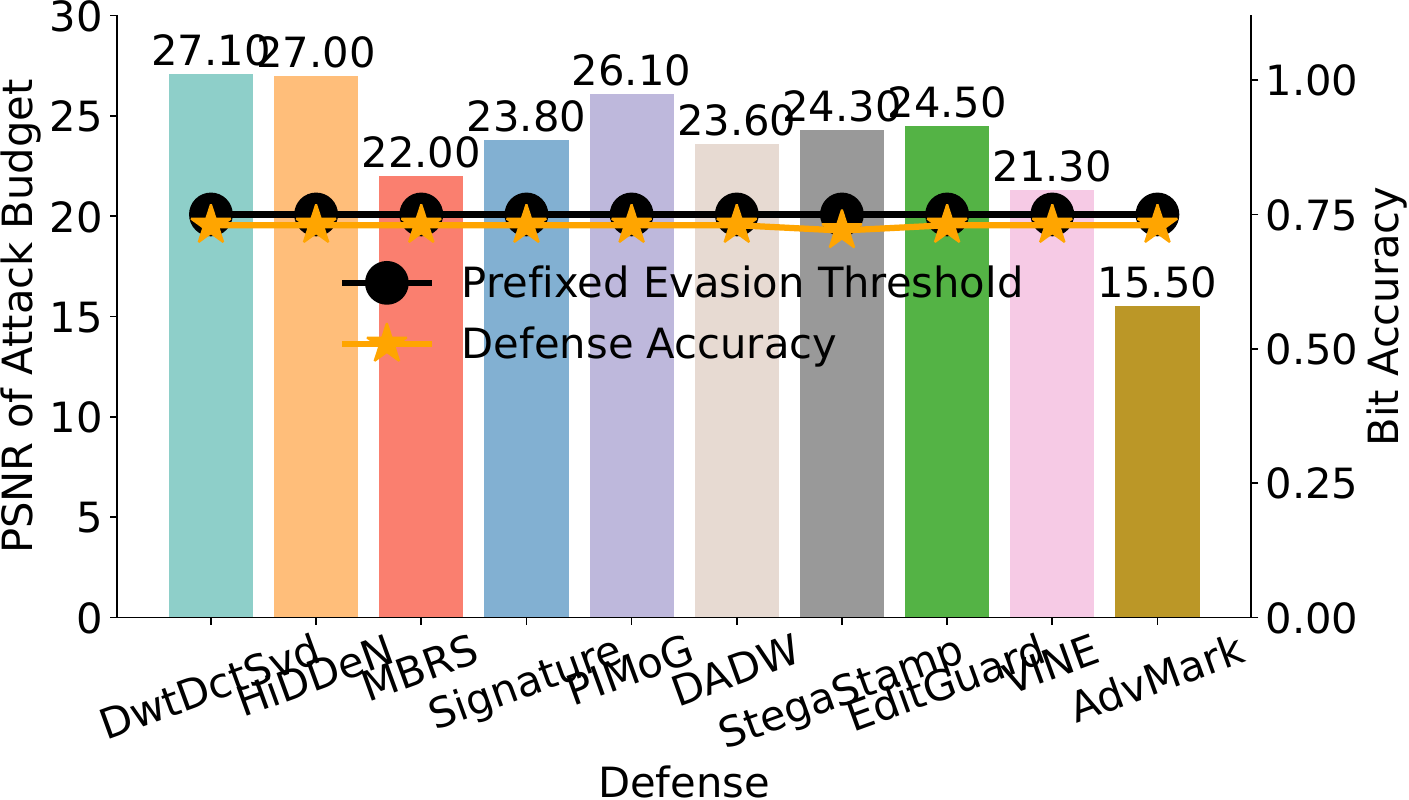}
        \caption{PSNR$\downarrow$ of Black-Q attacks.}
        \label{fig:surrogate_accuracy}
    \end{subfigure}
    \caption{Attack performance of Black-S and Black-Q.}
    \label{figure_6}
\end{figure}

\begin{figure}[]
    \centering
    \begin{subfigure}{0.48\linewidth}
        \centering
        \includegraphics[width=\linewidth]{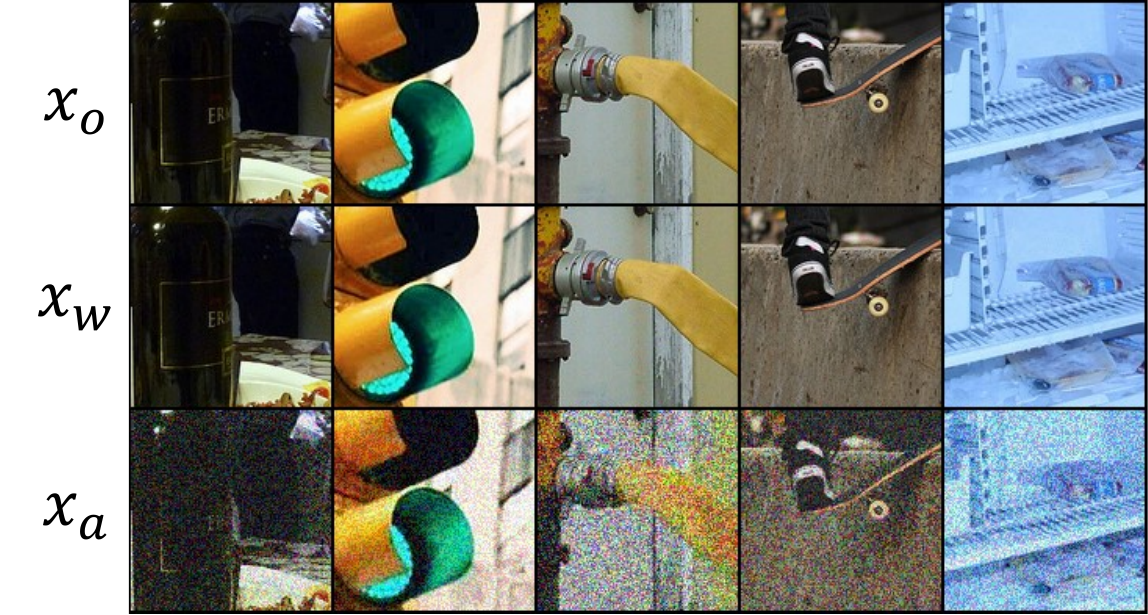}
        \caption{Image of Black-Q attacks.}
        \label{fig:surrogate_accuracy}
    \end{subfigure}
    \begin{subfigure}{0.48\linewidth}
        \centering
        \includegraphics[width=\linewidth]{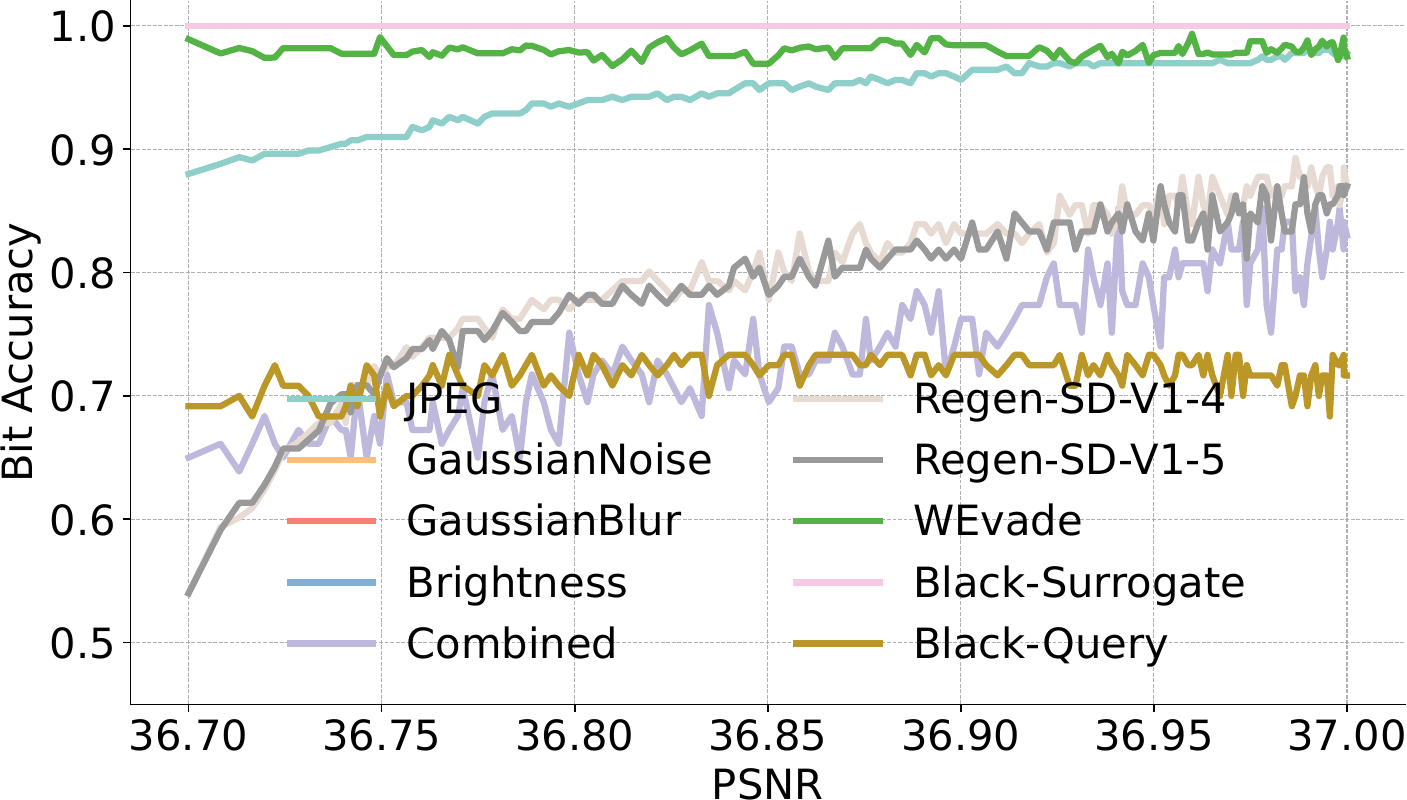}
        \caption{Accuracy$\uparrow$ in stage 2.}
        \label{fig:ba_wevade}
    \end{subfigure}%
    \caption{Example of Black-Q attack and accuracy optimization.}
    \label{figure_opt_attack}
\end{figure}

\subsection{Watermark Robustness}
\label{subsec_robustness}
Table \ref{table_acc} depicts the comprehensive robustness across all attack where ours maintains the highest clean accuracy.
For distortion and regeneration attacks: AdvMark outperforms the SOTA methods by up to 29\% against JPEG and 33\% against V1-4 attacks. This stems from our efficient attack optimization in Equ.~\ref{equ_La_2}.
In comparison, MBRS achieves high JPEG robustness but yields a low accuracy of 0.76 against combined distortions compared to 0.83.

For adversarial attacks, AdvMark outperforms baselines with 16\%-46\% accuracy improvement.
Note that DwtDctSvd is composed of non-differentiable transformations and henceforth is not applicable.
MBRS ranks the second with only 0.82 accuracy due to poor joint training.
Thanks to our fine-tuning in stage 1 where encoded images are moved into non-attackable regions, it mitigates Black-S attack by confusing surrogate models with fine-tuned images distribution.
As shown in Fig. \ref{figure_6} (a), we can find that lower validation accuracy generally aligns with higher bit accuracy (robustness) e.g., accuracy of 0.65 and 0.77 from AdvMark and MBRS correspond to both 1.00 in Table \ref{table_acc}.

For Black-Q attack, it initializes an image with random noise, and then iteratively optimizes the sample towards the target image but always guarantees successful evasion.
Therefore the bit accuracy is always below threshold $\tau'=0.75$ and only the perturbation budget differs.
We present the PSNR of the attacked image in Fig.~\ref{figure_6} (b).
As expected, AdvMark extracts a significant quality loss ($\sim$15 PSNR) from Black-Q attack, thanks to our image moving strategy.
We also present some examples of the Black-Q attacked images from AdvMark in Fig.~\ref{figure_opt_attack} (a).


\begin{table}
\centering
\caption{Ablation of AdvMark. AdvMark w/ $\lambda_{i_2}=7$ achieves better robustness but at the cost of more optimization overheads. AdvMark w/ different sizes and n still outperforms baselines.}
\resizebox{1.0\linewidth}{!}
{\begin{tabular}{c|c|ccccc|cc|ccc}
\toprule[1.6pt]
\multirow{2}{*}{\diagbox[]{Defense}{Attack}} & \multirow{2}{*}{PSNR} & \multicolumn{5}{c|}{Distortion}                                               & \multicolumn{2}{c|}{Regeneration} & \multicolumn{3}{c}{Adversarial}               \\ \cmidrule{3-12} 
                                             &                       & 1             & 2             & 3             & 4             & 1$\sim$4      & V1-4            & V1-5            & WEvade        & B-S           & B-Q           \\ \midrule
{AdvMark (Ours)}                      & \cellcolor[HTML]{ECF4FF}{37.0}                  & 0.99          & \cellcolor[HTML]{ECF4FF}{1.00} & \cellcolor[HTML]{ECF4FF}{1.00} & \cellcolor[HTML]{ECF4FF}{1.00} & 0.83          & 0.87            & 0.87            & \cellcolor[HTML]{ECF4FF}{0.98} & \cellcolor[HTML]{ECF4FF}{1.00} & \cellcolor[HTML]{ECF4FF}{0.73} \\
w/o Stage 1                                  & \colorbox[HTML]{FFE5E3}{34.7}                  & \cellcolor[HTML]{ECF4FF}{1.00} & 1.00          & 1.00          & 1.00          & \cellcolor[HTML]{ECF4FF}{0.85} & \cellcolor[HTML]{ECF4FF}{0.88}   & \cellcolor[HTML]{ECF4FF}{0.88}   & \colorbox[HTML]{FFE5E3}{0.50} & 1.00          & 0.73          \\
w/o Stage 2                                  & 36.7                  & \colorbox[HTML]{FFE5E3}{0.88} & 1.00          & 1.00          & 1.00          & \colorbox[HTML]{FFE5E3}{0.65} & \colorbox[HTML]{FFE5E3}{0.54}   & \colorbox[HTML]{FFE5E3}{0.54}   & 0.99          & 1.00          & \colorbox[HTML]{FFE5E3}{0.69} \\ \midrule
$iter_E$=5                                   & 36.5                  & 0.99          & 1.00          & 1.00          & 1.00          & 0.82          & 0.87            & 0.87            & 0.91          & 1.00          & 0.73          \\
$iter_E$=20                                  & 36.1                  & 0.90          & 1.00          & 1.00          & 1.00          & 0.75          & 0.85            & 0.85            & 1.00          & 1.00          & 0.73          \\
$\lambda_{i_2}=3$                            & 36.4                  & 1.00          & 1.00          & 1.00          & 1.00          & 0.83          & 0.88            & 0.88            & 0.98          & 1.00          & 0.73          \\
$\lambda_{i_2}=7$                            & 37.2                  & 1.00          & 1.00          & 1.00          & 1.00          & 0.85          & 0.89            & 0.89            & 0.99          & 1.00          & 0.73          \\ \midrule
size=128,n=48                                & 37.2                  & 1.00          & 1.00          & 1.00          & 1.00          & 0.84          & 0.87            & 0.88            & 0.98          & 1.00          & 0.73          \\
size=256,n=64                                & 38.4                  & 1.00          & 1.00          & 1.00          & 1.00          & 0.89          & 0.92            & 0.92            & 1.00          & 1.00          & 0.73          \\
size=256,n=100                               & 38.9                  & 1.00          & 1.00          & 1.00          & 1.00          & 0.89          & 0.93            & 0.93            & 1.00          & 1.00          & 0.73          \\ \midrule[1.6pt]
\end{tabular}}
\label{table_ablation_advmark}
\end{table}

\begin{figure}
    \centering
    \begin{subfigure}{0.48\linewidth}
        \centering
        \includegraphics[width=\linewidth]{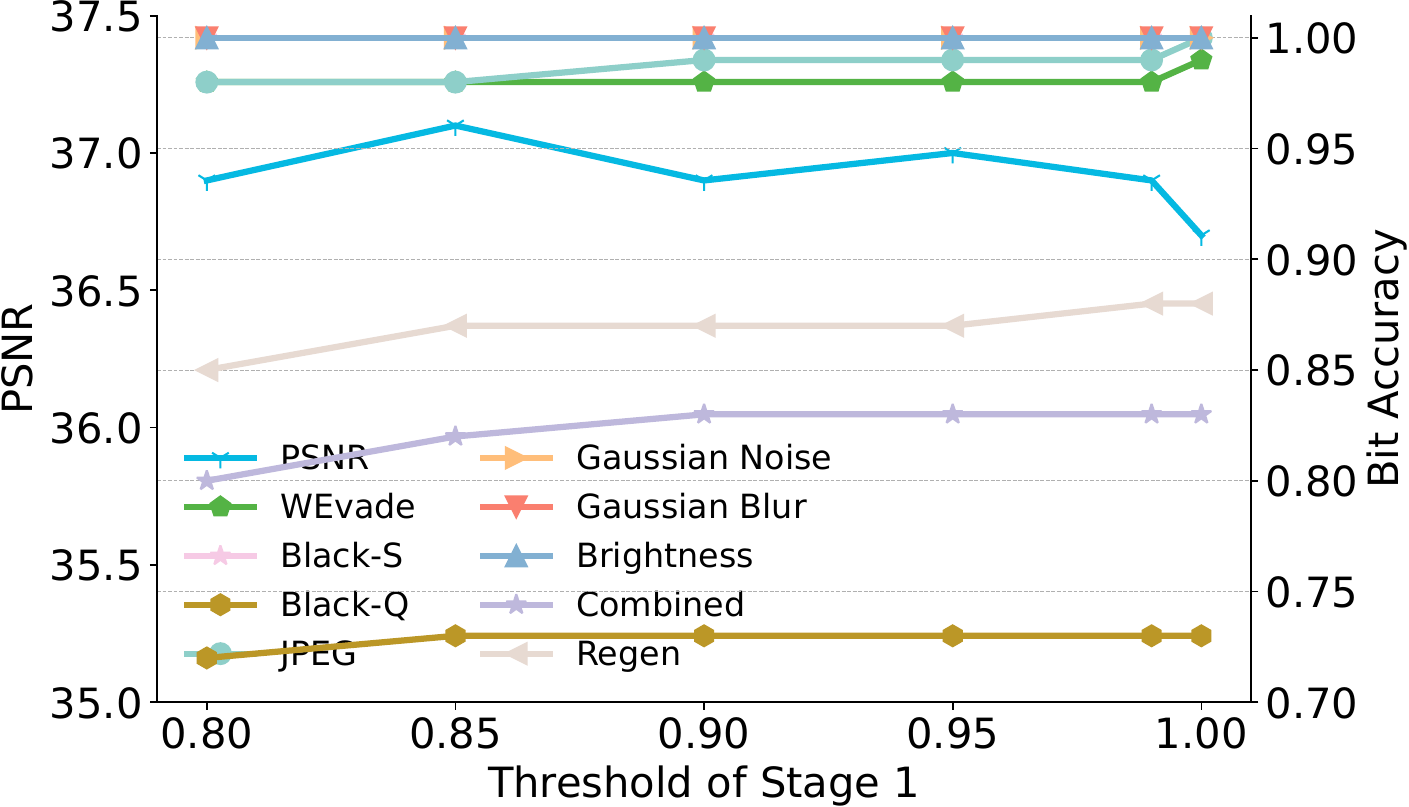}
        \caption{Stage 1}
        \label{fig:time_overhead}
    \end{subfigure}%
    \begin{subfigure}{0.48\linewidth}
        \centering
        \includegraphics[width=\linewidth]{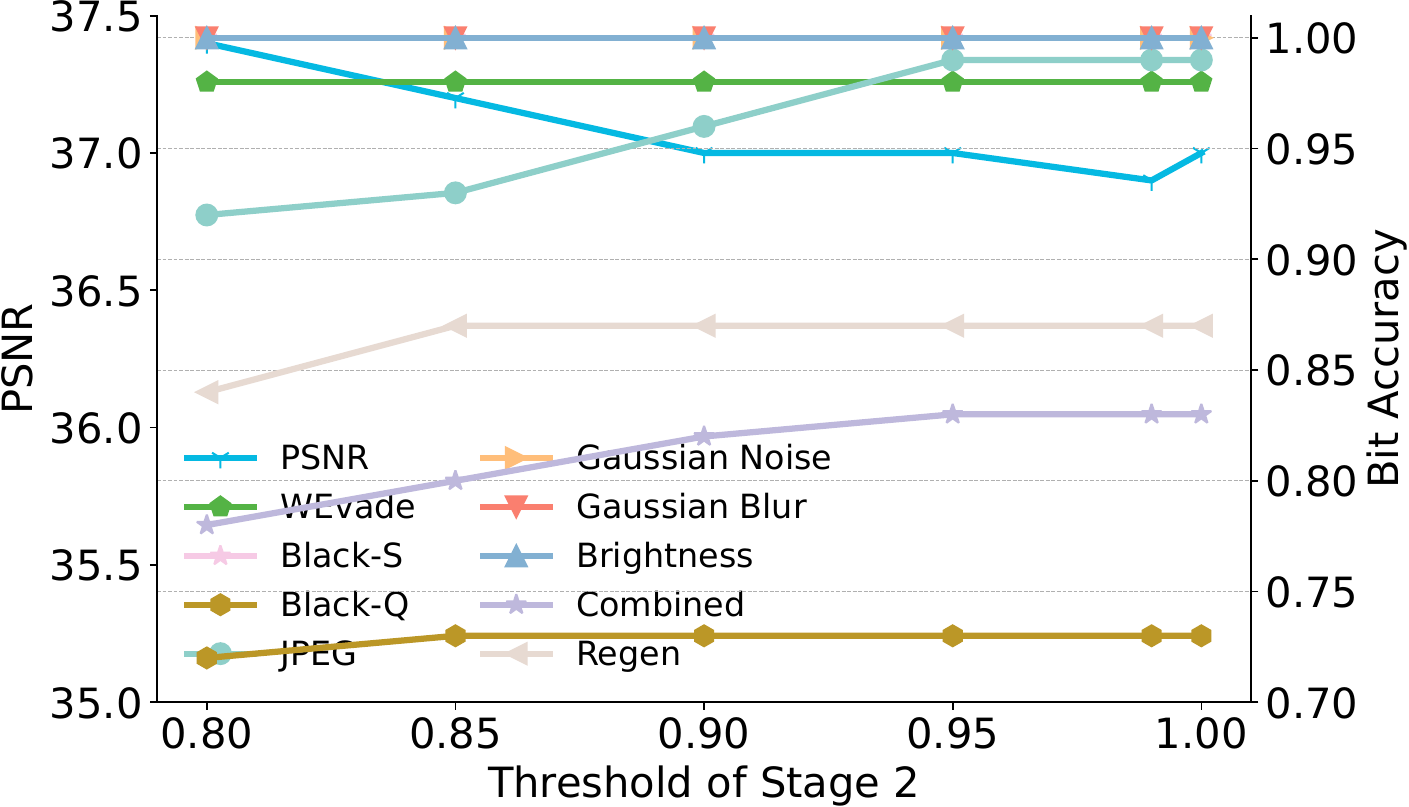}
        \caption{Stage 2}
        \label{fig:memory_overhead}
    \end{subfigure}
    \caption{PSNR $\uparrow$ and accuracy $\uparrow$ for different thresholds.}
    \label{figure_threshold}
\end{figure}

\begin{figure*}[]
    \centering
    \begin{subfigure}{0.24\linewidth}
        \centering
        \includegraphics[width=\linewidth]{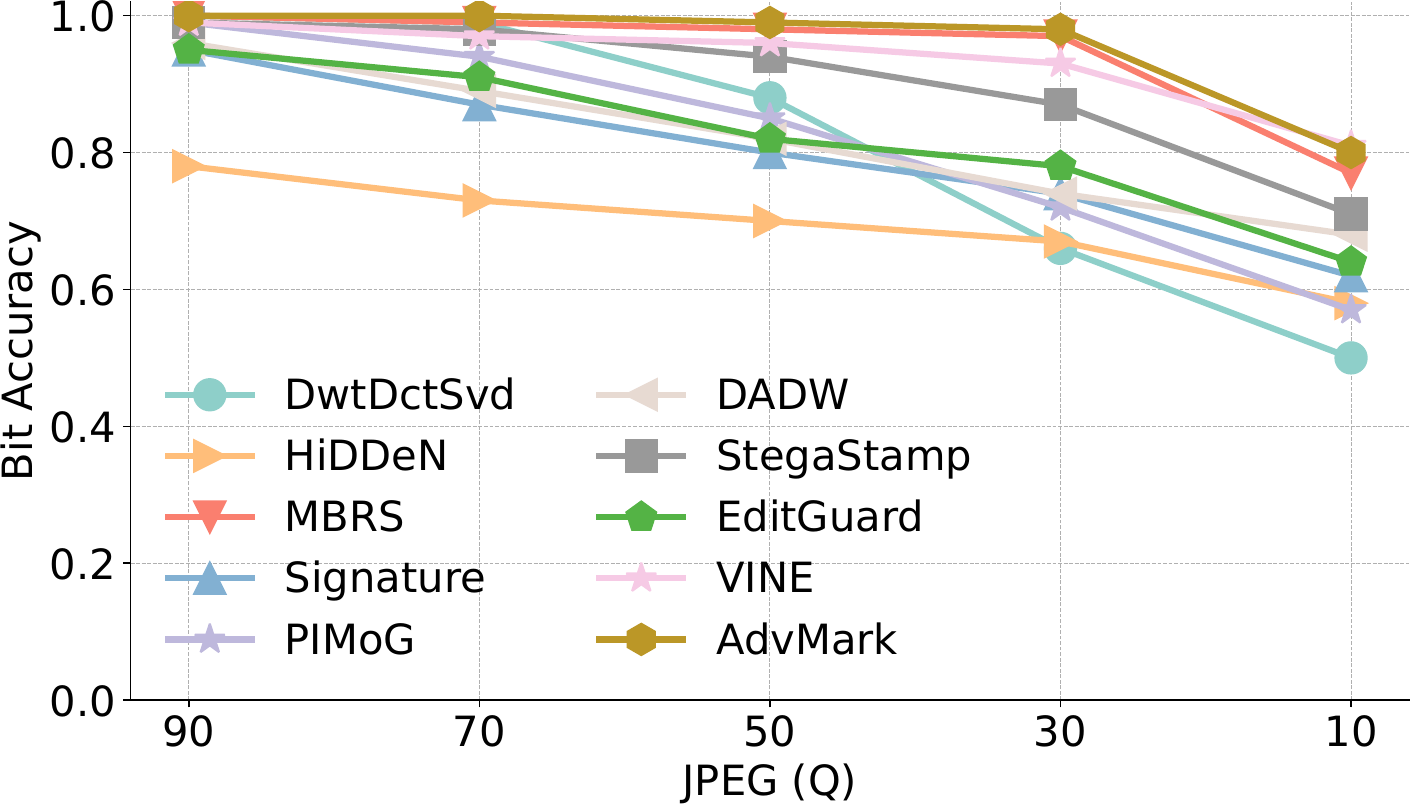}
        \caption{JPEG (Train Q=50)}
        \label{fig:ba_wevade}
    \end{subfigure}
    \begin{subfigure}{0.24\linewidth}
        \centering
        \includegraphics[width=\linewidth]{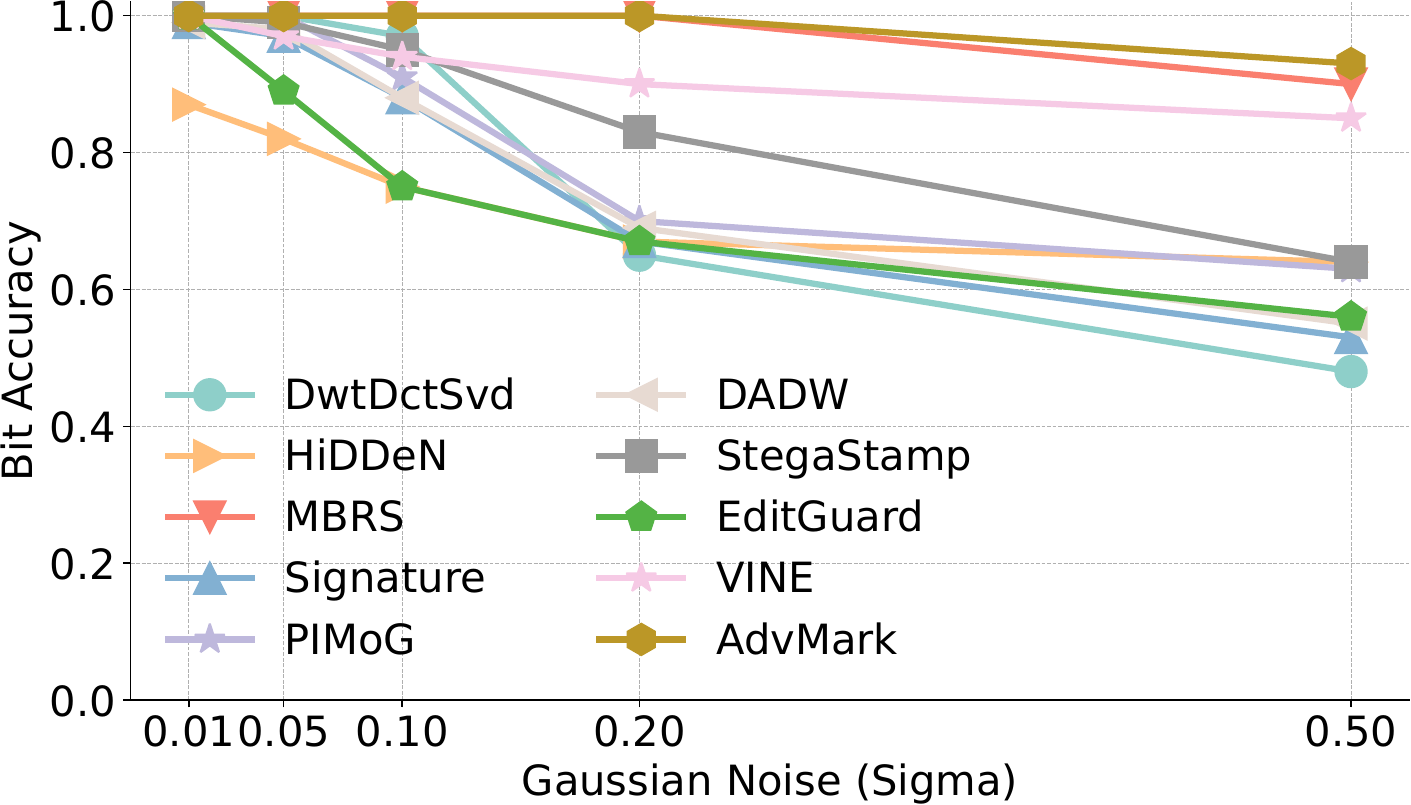}
        \caption{GaussianNoise (Train $\sigma$=0.1)}
        \label{fig:ba_wevade}
    \end{subfigure}
    \begin{subfigure}{0.24\linewidth}
        \centering
        \includegraphics[width=\linewidth]{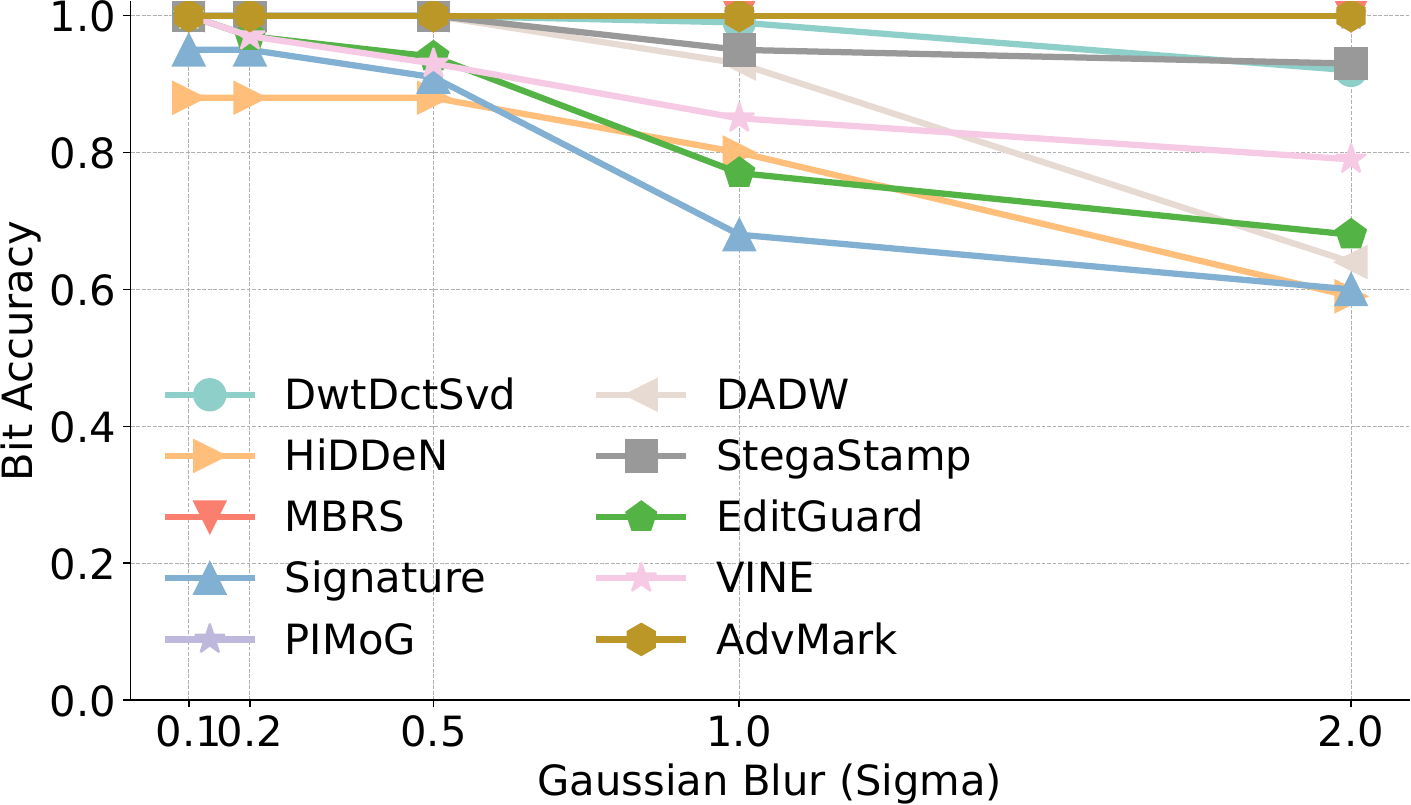}
        \caption{GaussianBlur (Train $\sigma$=0.5)}
        \label{fig:ba_wevade}
    \end{subfigure}
    \begin{subfigure}{0.24\linewidth}
        \centering
        \includegraphics[width=\linewidth]{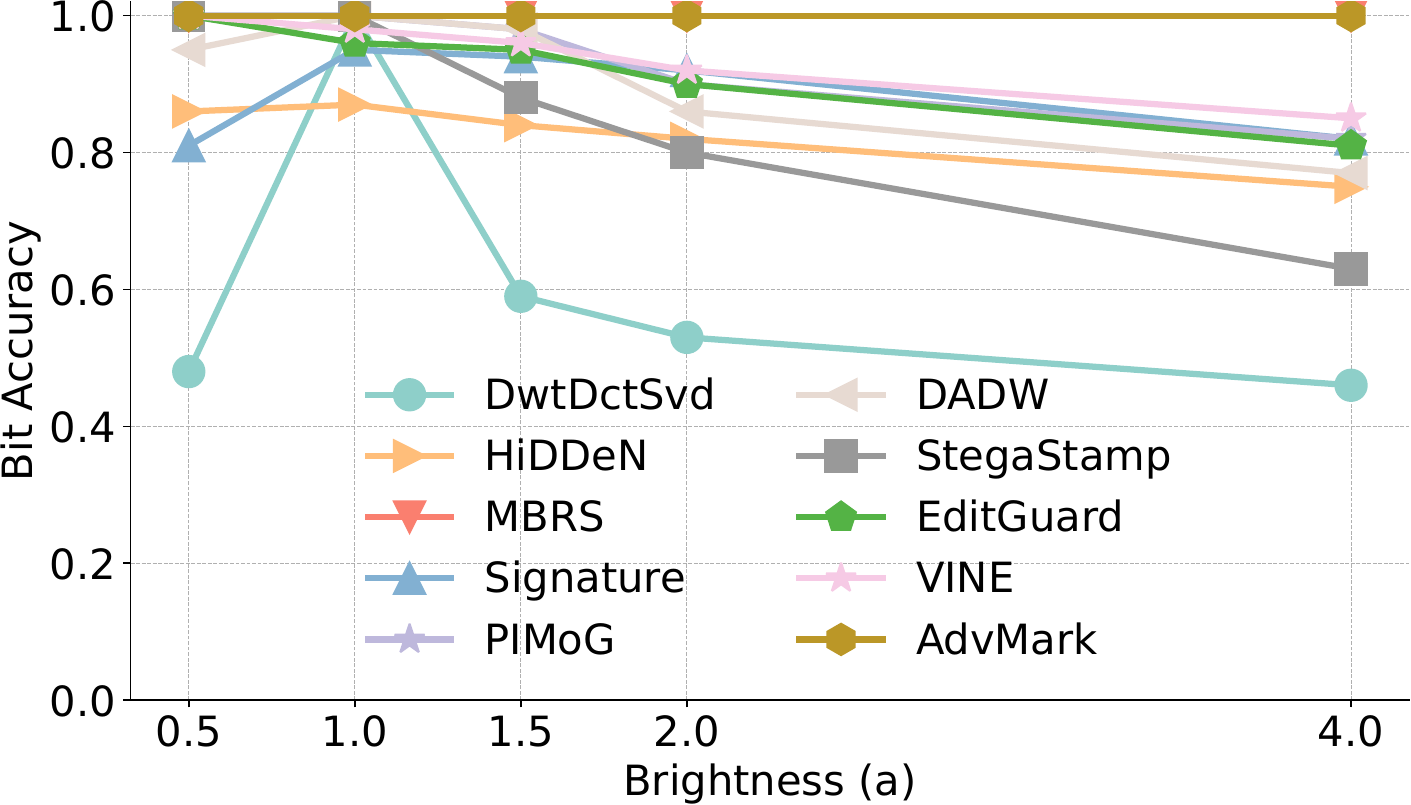}
        \caption{Brightness (Train a=1.5)}
        \label{fig:ba_wevade}
    \end{subfigure}
    \caption{Bit accuracy $\uparrow$ of watermarking methods against distortion attacks with different testing parameters.}
    \label{figure_14}
\end{figure*}

\subsection{Ablation Study}
\label{subsec_ablation}
\paragraph{\textbf{Effectiveness of two stages}}
As shown in Table \ref{table_ablation_advmark}.
We first skip stage 1 and replace the fine-tuned model with original MBRS.
Separately we also skip stage 2 to evaluate the robustness against distortion and regeneration attacks.
As expected, without stage 1, image quality degrades and AdvMark fails to defend against WEvade without encoder fine-tuning.
Without stage 2, AdvMark only enhances adversarial robustness and deteriorates distortion robustness, i.e. only 0.88 and 0.65 accuracy against JPEG and combined attacks, which implies the necessity of image optimization to restore robustness balance.

To better understand how stage 2 preserves previous robustness, we present the bit accuracy change during optimization in Fig.~\ref{figure_opt_attack} (b).
We can find that stage 1 of AdvMark generates high quality images (PSNR=36.7) but with poor robustness.
In stage two's optimization, AdvMark constantly enhances overall accuracy with slight quality improvement, while also maintaining adversarial robustness, thanks to our constrained image loss $l(x_{w_2},x_{w_1})$ in Equ.~\ref{equ_Li_2}.

\paragraph{\textbf{Parameter impact}}
We retrain AdvMark with different encoder fine-tuning iterations $iter_E$ in stage 1 and image loss weight $\lambda_{i_2}$ in stage 2.
In Table.~\ref{table_ablation_advmark}, regarding $iter_E$, we find that fewer fine-tuning iterations yield inferior adversarial robustness (0.91 WEvade accuracy) and lower visual quality since the encoder is not fully optimized.
However, high iterations tend to overfit adversarial attacks, which in turn degrades image quality (PSNR budget $p$=36) to gain JPEG and regeneration attack robustness.
Regarding $\lambda_{i_2}$, generally high image weight guarantees the optimization is fully explored until PSNR quality budget, therefore AdvMark achieves higher image quality with similar robustness at the cost of more computation overhead in stage 2.

\begin{table}
\centering
\caption{Ablation study of baselines. We present watermarking PSNR $\uparrow$ and robustness accuracy $\uparrow$. Baselines with stage 2 are enhanced, demonstrating our optimization effectiveness.}
\resizebox{\linewidth}{!}
{\begin{tabular}{c|c|ccccc|cc|ccc}
\toprule[1.6pt]
\multirow{2}{*}{\diagbox[]{Defense}{Attack}} & \multirow{2}{*}{PSNR} & \multicolumn{5}{c|}{Distortion}                                               & \multicolumn{2}{c|}{Regeneration} & \multicolumn{3}{c}{Adversarial}               \\ \cmidrule{3-12} 
                                             &                       & 1             & 2             & 3             & 4             & 1$\sim$4      & V1-4            & V1-5            & WEvade        & B-S           & B-Q           \\ \midrule
HiDDeN w/ Stage 2                            & 30.0                  & 0.88          & 0.98          & 1.00          & 1.00          & 0.77          & 0.73            & 0.73            & 0.54          & 0.77          & 0.73          \\
Stable Signature w/ Stage 2                  & 30.7                  & 0.92          & 1.00          & 1.00          & 1.00          & 0.79          & 0.78            & 0.79            & 0.76          & 1.00          & 0.73          \\
PIMoG w/ Stage 2                             & 36.4                  & 0.96          & 1.00          & 1.00          & 1.00          & 0.81          & 0.81            & 0.81            & 0.74          & 1.00          & 0.72          \\ \midrule
{AdvMark+HiDDeN (Ours)}               & \cellcolor[HTML]{ECF4FF}{30.7}         & \cellcolor[HTML]{ECF4FF}{0.85} & \cellcolor[HTML]{ECF4FF}{1.00} & \cellcolor[HTML]{ECF4FF}{1.00} & \cellcolor[HTML]{ECF4FF}{1.00} & \cellcolor[HTML]{ECF4FF}{0.75} & \cellcolor[HTML]{ECF4FF}{0.80}   & \cellcolor[HTML]{ECF4FF}{0.80}   & \cellcolor[HTML]{ECF4FF}{0.91} & \cellcolor[HTML]{ECF4FF}{1.00} & \cellcolor[HTML]{ECF4FF}{0.73} \\
HiDDeN                                       & \colorbox[HTML]{FFE5E3}{30.3}         & \colorbox[HTML]{FFE5E3}{0.70} & \colorbox[HTML]{FFE5E3}{0.75} & \colorbox[HTML]{FFE5E3}{0.87} & \colorbox[HTML]{FFE5E3}{0.85} & \colorbox[HTML]{FFE5E3}{0.64} & \colorbox[HTML]{FFE5E3}{0.57}   & \colorbox[HTML]{FFE5E3}{0.56}   & \colorbox[HTML]{FFE5E3}{0.57} & \colorbox[HTML]{FFE5E3}{0.68} & \colorbox[HTML]{FFE5E3}{0.73} \\ \midrule
MBRS, S=0.6                                  & 38.0                  & 0.87          & 0.98          & 0.99          & 0.98          & 0.68          & 0.60            & 0.61            & 0.72          & 1.00          & 0.72          \\
MBRS, S=1.0                                  & 32.1                  & 0.98          & 1.00          & 1.00          & 1.00          & 0.76          & 0.70            & 0.70            & 0.82          & 1.00          & 0.73          \\
MBRS, S=2.0                                  & 28.1                  & 1.00          & 1.00          & 1.00          & 1.00          & 0.83          & 0.78            & 0.78            & 0.89          & 1.00          & 0.73          \\ \midrule[1.6pt]
\end{tabular}}
\label{table_ablation_baseline}
\end{table}

To demonstrate the generalization of AdvMark, we also retrain with different image size and message length $n$ in Table \ref{table_ablation_advmark}, which aligns with other baselines in Table \ref{table_psnr}.
We can find that regardless of the parameters, AdvMark still outperforms with the highest quality and robustness, e.g. AdvMark (size=256, n=64, PSNR=38.4) surpasses StegaStamp with the same setting.

To demonstrate the impact of two thresholds $\tau_1$ and $\tau_2$, we present the PSNR and robustness changes in Fig. \ref{figure_threshold}.
Generally, when threshold 1 is higher than 0.98, it degrades WEvade robustness due to insufficient training from optional decoder.
With lower threshold 2, there is a risk of insufficient optimization, e.g. lower JPEG and Regenerative robustness.
In the end, we empirically choose both thresholds as 0.95 for best robustness.

\paragraph{\textbf{Ablation on baselines}}
To demonstrate the effectiveness of stage 2 on baselines, we apply image fine-tuning directly after the baseline models (HiDDeN, Stable Signature and PIMoG) and present the results in Table \ref{table_ablation_baseline}.
It is solid that the robustness against distortion and regeneration attacks is enhanced, compared with the original performance in Table \ref{table_acc}, thanks to our direct optimization.

To demonstrate the effectiveness of AdvMark on baselines, we apply the two-stage enhancement on another base model HiDDeN in Table.~\ref{table_ablation_baseline}.
It is as expected that AdvMark+HiDDeN significantly improves comprehensive robustness against all 10 attacks with only 0.4dB PSNR improvement.
This is because HiDDeN itself is extremely vulnerable to even simple distortion attacks, therefore AdvMark exchanges more quality for higher robustness as much as up to 34\% accuracy improvement.

To evaluate whether we can exchange robustness of image quality, we fine-tune MBRS with joint adversarial training under different strength factors S=0.6, 1.0, 2.0 in Table.~\ref{table_ablation_baseline}.
It depicts that MBRS with higher S yields lower visual quality with higher robustness.
Nonetheless, AdvMark still outperforms the S=2.0 baseline with 9\% accuracy improvement against regeneration and adversarial attacks, let alone our significantly high image quality.
MBRS with S=0.6 shows higher quality yet exhibits poor robustness against all three kinds of attacks.
As a result, solely relying on adjusted strength factor cannot suffice compared with AdvMark.
\begin{table}
\centering
\caption{Performance of geometric and advanced attacks. We present the robustness accuracy $\uparrow$. Although MBRS achieves similar results as AdvMark, it fails to defend against regeneration and adversarial attack, as shown in Table \ref{table_acc}.}
\resizebox{\linewidth}{!}
{\begin{tabular}{c|cccccc|cc}
\toprule[1.6pt]
\multirow{2}{*}{\diagbox[]{Defense}{Attack}} & \multicolumn{6}{c|}{Geometry Attack}                                                          & \multicolumn{2}{c}{Advanced Combined} \\ \cmidrule{2-9} 
                                             & Crop          & Resize        & Dropout       & SaltPepper    & Rotation      & Hue           & Reg+Adv           & Adv+Reg           \\ \midrule
HiDDeN                                       & \cellcolor[HTML]{FFE5E3}{0.86} & \cellcolor[HTML]{FFE5E3}{0.83} & \cellcolor[HTML]{FFE5E3}{0.90} & \cellcolor[HTML]{FFE5E3}{0.89} & \cellcolor[HTML]{FFE5E3}{0.85} & \cellcolor[HTML]{FFE5E3}{0.88} & 0.57              & 0.52              \\
DADW                                         & 0.91          & 0.85          & 0.96          & 0.95          & 0.90          & 0.93          & \cellcolor[HTML]{FFE5E3}{0.54}     & \cellcolor[HTML]{FFE5E3}{0.49}     \\
MBRS                                         & 0.97          & 1.00          & 1.00          & 1.00          & 0.94          & 1.00          & 0.70              & 0.65              \\
AdvMark                                      & \cellcolor[HTML]{ECF4FF}{1.00} & \cellcolor[HTML]{ECF4FF}{1.00} & \cellcolor[HTML]{ECF4FF}{1.00} & \cellcolor[HTML]{ECF4FF}{1.00} & \cellcolor[HTML]{ECF4FF}{1.00} & \cellcolor[HTML]{ECF4FF}{1.00} & \cellcolor[HTML]{ECF4FF}{0.87}     & \cellcolor[HTML]{ECF4FF}{0.81}     \\ \midrule[1.6pt]
\end{tabular}}
\label{table_ablation_distortion}
\end{table}

\paragraph{\textbf{More attack settings}}
To demonstrate adaptation to different attack strength, we present the bit accuracy against distortion attacks with different parameters in Fig.~\ref{figure_14}.
The results indicate that AdvMark still outperforms in all the scenarios, e.g. $\sim$0.8 and $\sim$0.6 accuracy against JPEG with Q=10 for AdvMark and other baselines.

Finally, we include more geometric (following DWSF \cite{guo2023practical}) and combined attacks in Table \ref{table_ablation_distortion}.
The parameters are Crop (p=0.035), Resize (r=0.7), Dropout(p=0.3), SaltPepper(p=0.1), Rotation(a=30$^{\circ}$) and Hue($\delta$=0.2).
In general, AdvMark still outperforms with all 1.0 accuracy with MBRS ranking the second, which validates the high robustness of the base model MBRS.
For combined attacks, Reg+Adv is basically the same as Reg alone due to significant regeneration quality degradation that breaks the adversarial budget.

\subsection{Overhead Analysis}
\begin{figure}
    \centering
    \begin{subfigure}{0.48\linewidth}
        \centering
        \includegraphics[width=\linewidth]{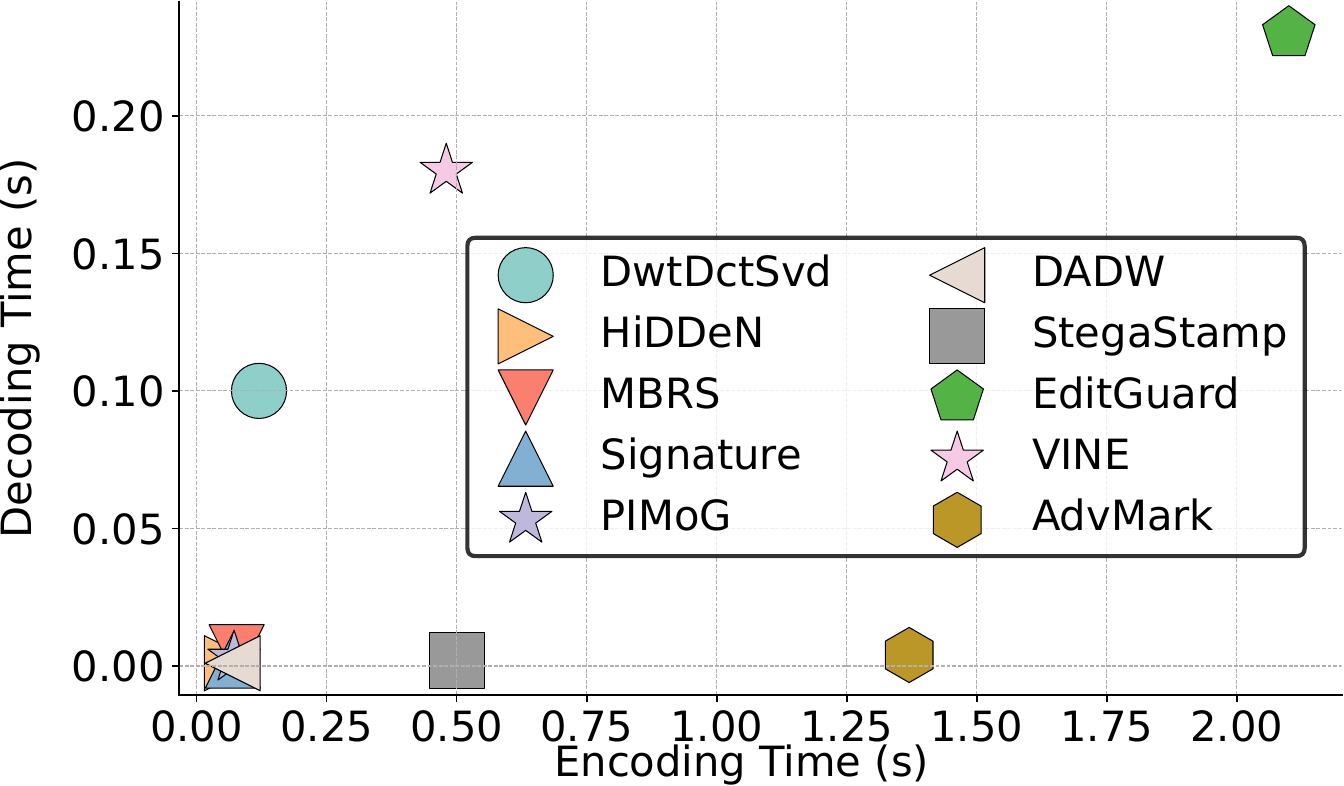}
        \caption{Time overhead $\downarrow$.}
        \label{fig:time_overhead}
    \end{subfigure}%
    \begin{subfigure}{0.48\linewidth}
        \centering
        \includegraphics[width=\linewidth]{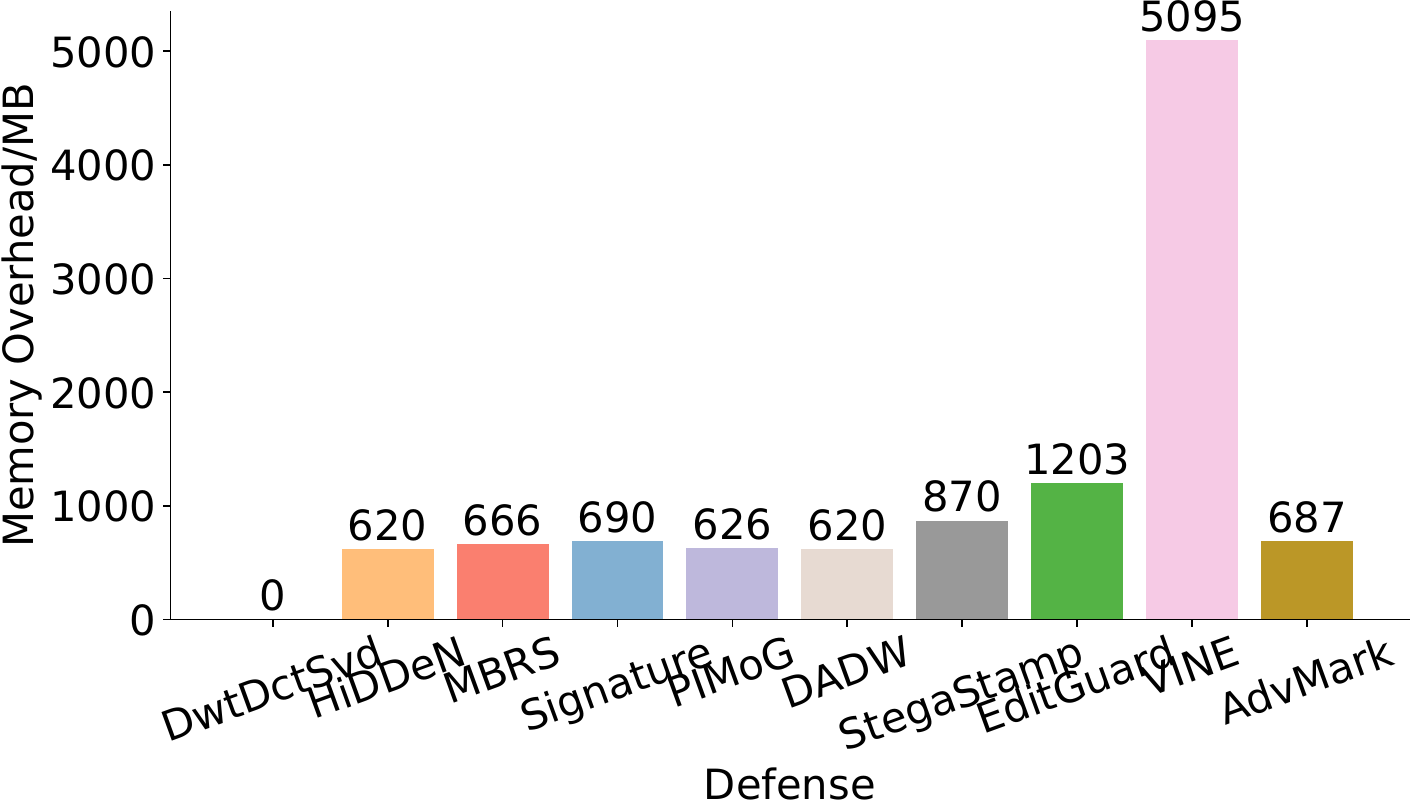}
        \caption{GPU Memory overhead $\downarrow$.}
        \label{fig:memory_overhead}
    \end{subfigure}
    \caption{Overhead comparison.}
    \label{figure_9}
\end{figure}

We also evaluate AdvMark in terms of time and GPU memory overhead.
For training overhead, AdvMark exhibits only O(1/2*N) and typical JAT requires O(N) to update both encoder and decoder, where N is the total iterations.
For inference, the results are presented in Fig.~\ref{figure_9}.
AdvMark achieves real time decoding and only incurs acceptable overhead from image optimization.
EditGuard utilizes an inverse neural network to encode both secret images and bits simultaneously, which leads to longer processing time.
While VINE requires high GPU memory due to heavy diffusion model.
In summary, AdvMark can achieve SOTA performance within reasonable overhead.
\section{Conclusion}
\label{sec_conclusion}
In this paper, we first reveal the challenges of typical joint optimization against distortion, regeneration and adversarial attacks.
Built upon the insights, we propose AdvMark which decouples a two-stage defense method.
In stage one, we propose a defender tailored adversarial attack, along with a novel conditional training paradigm to mainly fine-tune the encoder to maintain clean accuracy.
While in stage two, we directly optimize the encoded image to address other attacks.
We propose a novel constrained image loss to preserve adversarial robustness with verifiable theoretical guarantees.
To further guarantee visual quality, we improve upon the PGD optimization with a metric-aware early-stop.
Extensive experiments demonstrate our effectiveness.
\section{Acknowledgements}
This work was supported by Key Laboratory of Data Intelligence, Beijing.
{
    \small
    \bibliographystyle{ieeenat_fullname}
    \bibliography{main}
}


\end{document}